\documentclass{article}

\usepackage[utf8]{inputenc} 
\usepackage[T1]{fontenc}    
\usepackage[hidelinks]{hyperref}       
\usepackage{url}            
\usepackage{booktabs}       
\usepackage{amsfonts}       
\usepackage{nicefrac}       
\usepackage{microtype}      

\usepackage{enumitem}

\PassOptionsToPackage{numbers, compress}{natbib}
\usepackage[preprint]{nips_2018}

\usepackage{amsmath}
\usepackage{amsthm}
\usepackage{amssymb}
\usepackage{graphicx}
\usepackage{wrapfig}
\usepackage{xspace}
\usepackage{extarrows}
\usepackage{xfrac}
\usepackage[most]{tcolorbox}
\usepackage{sidecap}        \sidecaptionvpos{figure}{t}

\newtheorem{proposition}{Proposition}
\newtheorem{definition}{Definition}
\theoremstyle{definition}

\newtheorem{example}{Example}

\newtcolorbox{exbox}{colback=gray!15,colframe=black,breakable,enhanced}

\numberwithin{equation}{section}

\newcommand*{\ie}{i.e.\@\xspace}

\makeatletter
\newcommand*{\etc}{%
    \@ifnextchar{.}%
        {etc}%
        {etc.\@\xspace}%
}
\makeatother

\newcommand{\bs}[1]{\boldsymbol{#1}} 
\renewcommand{\vec}[1]{\mathbf{#1}}
\newcommand{\mat}[1]{\mathbf{#1}}
\newcommand{\x}[0]{\vec{x}}

\newcommand{\z}[0]{\vec{z}}
\newcommand{\J}[0]{\mat{J}}
\newcommand{\Mz}[0]{\mat{M}_{\z}}
\newcommand{\dif}[1]{\mathrm{d}#1} 
\newcommand{\dist}[0]{\mathsf{dist}} 
\newcommand{\inner}[2]{\langle#1,#2\rangle} 
\newcommand{\vectorize}[1]{\text{vec}\!\left[ #1\right]} 
\newcommand{\var}[1]{\mathsf{var}\left[ #1 \right]} 
\newcommand{\cov}[1]{\mathsf{cov}\left[ #1 \right]} 
\newcommand{\T}[0]{^{\top}}
\newcommand{\away}{\xlongrightarrow{\mathrm{away}}}
\newcommand{\near}{\xlongrightarrow{\mathrm{near}}}
\DeclareMathOperator{\trace}{tr}
\newcommand{\xtx}[1]{#1\T \mkern-1.5mu\relax #1} 

\def\R{\mathbb{R}}  
\def\E{\mathbb{E}}  
\def\N{\mathcal{N}} 
\def\M{\mathcal{M}} 
\def\Z{\mathcal{Z}} 
\def\X{\mathcal{X}} 

\begin{document}

\title{\vspace{-1mm}Only Bayes should learn a manifold\\ {\large (on the estimation of differential geometric structure from data)}\vspace{-2mm}}

%

\author{
  S{\o}ren Hauberg \\
  Section for Cognitive Systems\\
  Technical University of Denmark\\
  \texttt{sohau@dtu.dk}
}

\maketitle

\begin{abstract}
  We investigate learning of the differential geometric structure of a data manifold
  embedded in a high-dimensional Euclidean space. We first analyze
  kernel-based algorithms and show that under the usual regularizations,
  non-probabilistic methods cannot recover the differential geometric structure,
  but instead find mostly linear manifolds or spaces equipped with teleports.
  We repeat the analysis for probabilistic methods and show that they naturally recover the geometric structure.
  Fully exploiting this
  structure, however, requires the development of stochastic extensions to classic
  Riemannian geometry. We take early steps in that regard.
  Finally, we partly extend the analysis to models based on neural networks,
  thereby highlighting geometric and probabilistic shortcomings
  of current deep generative models.
\end{abstract}

\begin{exbox}
  \begin{center}
  \resizebox{\textwidth}{!}{\small
  Comments on this document are gratefully accepted at \url{sohau@dtu.dk}.
  This is the $2^{\text{nd}}$ revision. }
  \end{center}
\end{exbox}

\section{Motivation and background}
  \emph{Manifold learning} aim to learn a low-dimensional representation of data
  that reflect the intrinsic structure of data. Spectral methods
  seek a low-dimensional embedding of high-dimensional data that preserve
  certain aspects of the data. This includes  methods
  such as \emph{Isomap} \cite{isomap}, \emph{Locally linear embeddings} \cite{lle},
  \emph{Laplacian eigenmaps} \cite{Belkin:2003:LED} and more \cite{Scholkopf99:kernelprincipal, Donoho5591}.
  Probabilistic methods often
  view the data manifold as governed by a latent variable along
  with a generative model that describe how the latent manifold is to be embedded in 
  the data space. The common theme is the quest for a low-dimensional representation
  that faithfully capture the data. 
  
  Ideally, we want an \emph{operational representation}, \ie we want to be able
  to make mathematically meaningful calculations with respect to the learned
  representation. For quantitative data analysis in the learned representation,
  a reasonable set of supported ``operations'' at least include:
  \begin{itemize}[noitemsep,nolistsep]
    \item \textbf{Interpolation:} given two points, a natural unique interpolating
      curve that follow the manifold should exist.
    \item \textbf{Distances:} the distance between two points should be well-defined
      and 
      reflect the amount of energy required to transform one point to another.
    \item \textbf{Measure:} the representation should be equipped with a measure
      under which integration is well-defined for all points on the manifold.
  \end{itemize}
  Depending on which analysis is to be performed in the new representation, one
  may focus on different operations. Even if the above operations should be considered
  elementary, most manifold learning schemes do not support any of these.

  \textbf{Embedding methods} seek a low-dimensional embedding $\z_{1:N} = \{\z_1, \ldots, \z_N\}$
  of the data $\x_{1:N}$. These methods fundamentally only describe
  the data manifold at the points where data is observed and nowhere else.
  As such, the low-dimensional embedding space is only well-defined at $\z_{1:N}$.
  It is common to treat the low-dimensional embedding space as being Euclidean,
  but this is generally a \emph{post hoc} assumption with limited grounding in the
  embedding method. Fundamentally, the learned representation space is
  a discrete space that does not lend itself to continuous interpolations.
  Likewise, the most natural measure will only assign
  mass to the points $\z_{1:N}$, and any associated  distribution
  will be discrete. This is too limited to be considered an operational representation.
  
\begin{SCfigure}
    \includegraphics[width=0.3\textwidth]{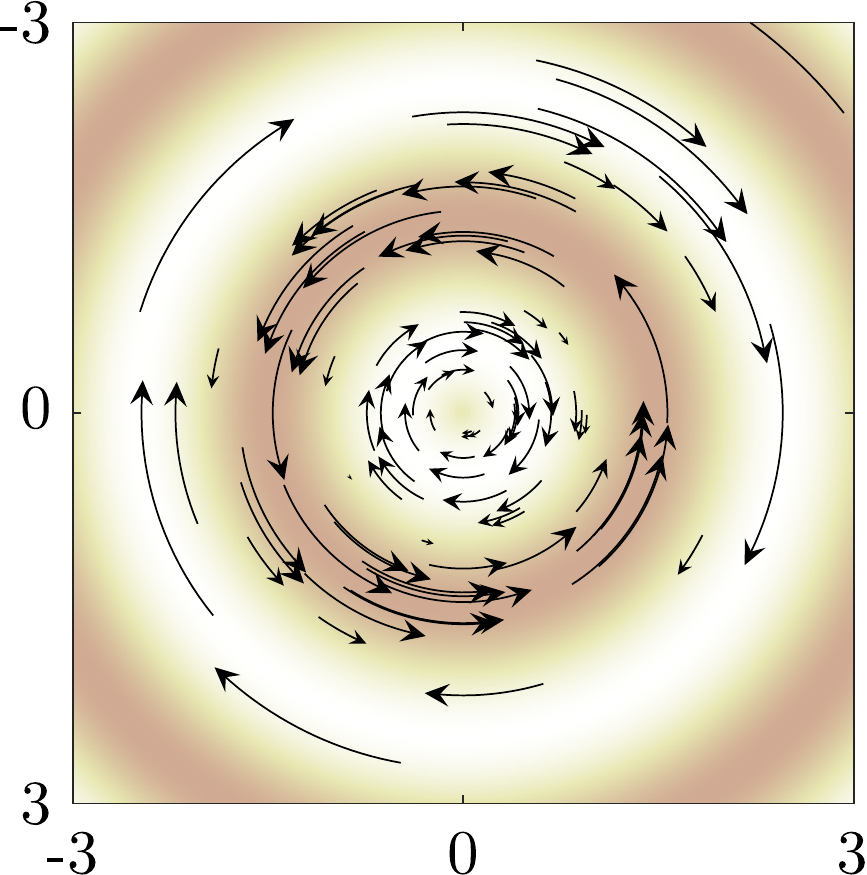}
    \includegraphics[width=0.3\textwidth]{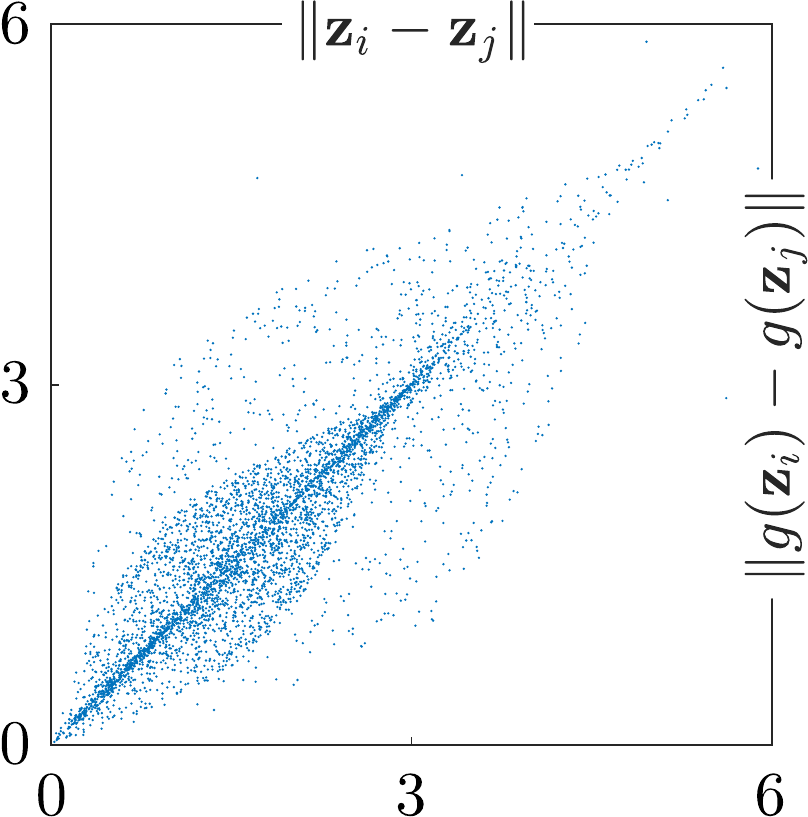}
    \caption{\protect\rule{0ex}{0ex} Reparametrizations illustrated.
      The left panel shows a ``swirling'' transformation of $\Z$ with the property
      that a Gaussian variable with zero mean and unit covariance, will have
      the same distribution after a reparametrization. The right panel shows
      pair-wise distances between points before and after reparametrization;
      evidently the geometry of $\Z$ is sensitive to reparametrizations.}
    \label{fig:swirl}    
  \end{SCfigure}

  \textbf{Generative models} estimate a set of low-dimensional latent
  variables $\z_{1:N}$ along with a suitable mapping $f: \Z \rightarrow \X$
  such that $f(\z) \approx \x$. It is, again, common to treat the latent
  space $\Z$ as being Euclidean. However, this assumption easily lead to 
  arbitrariness. 
  As an example, consider the \emph{variational autoencoder (VAE)} \cite{kingma:iclr:2014, rezende2014stochastic}, which seek
  a representation in which $\z_{1:N}$ follow a unit Gaussian distribution.
  Now consider the transformation
  \begin{align}
    g(\z) = \mat{R}_{\theta} \z,
  \end{align}
  where $\mat{R}_{\theta}$ is a linear transformation that rotate points by
  $\theta(\z) = \sin(\pi \|\z\|)$. This is a smooth invertible transformation
  with the property that
  \begin{align}
    \z \sim \N(\vec{0}, \mat{I}) \quad \Rightarrow \quad g(\z) \sim \N(\vec{0}, \mat{I}).
  \end{align}
  Figure~\ref{fig:swirl} illustrate this transformation. If the latent variables
  $\z_{1:N}$ and the generator $f$ is an optimal VAE, then $g(\z_{1:N})$ and
  $f \circ g^{-1}$ is equally optimal. Yet, the latent spaces $\Z$ and $g(\Z)$
  are quite different; Fig.~\ref{fig:swirl} shows the Euclidean distances between
  $\z_n$ and $g(\z_n)$ for samples drawn from a unit Gaussian. Clearly,
  the transformed latent space is significantly different from the original
  space. As the VAE provides no guarantees as to which latent space is recovered,
  we must be careful when relying on the Euclidean latent space: distances
  between points are effectively arbitrary and as are straight-line interpolations.
  Any analysis relying on vector operations in the latent space are, thus, arbitrary
  and positive result should be viewed either as pure luck with little mathematical
  grounding, or due to unspecified aspects of the model. Ideally, we
  want a representation space that is invariant to such transformations,
  but current models do not deliver. 

  \textbf{In this paper}, we consider models where the representation space $\Z$
  is learned jointly with a smooth mapping $f: \Z \rightarrow \X$, such that
  $\Z$ is naturally endowed with a Riemannian metric (Sec.~\ref{sec:primer}).
  This gives well-defined interpolants, distances 
  and a natural measure. We contribute a detailed analysis of the case
  where $f$ is estimated by a kernel method (Sec.~\ref{sec:kernels}), and show that even in the case of infinite
  noise-free data a non-probabilistic estimate of $f$ cannot recover the true
  Riemannian structure of $\Z$. In contrast, we show that probabilistic estimates of $f$
  can recover the true Riemannian structure (Sec.~\ref{sec:probabilistic}). Fully exploiting this
  structure, however, require the development of Bayesian extensions to classic
  differential geometry (Sec.~\ref{sec:bayes_geom}); we contribute elementary results in that regard, but
  many questions remain open. Finally, we partly extend our analysis to the case
  where $f$ is a neural network and demonstrate that current deep generative
  models are lacking elementary properties before they can learn the
  Riemannian structure of data manifolds (Sec.~\ref{sec:deep}).
  Our key finding is that uncertainty quantification is a prerequisite for learning an
  operational representation as the usual smoothness regularization introduce a harmful bias.
  
  \textbf{Notation.} We let $\Z$ denote the $d$-dimensional
  \emph{representation} or \emph{latent space}, which is learned from data
  in the \emph{observation space} $\X \equiv \R^D$. Topologically, the latent space
  $\Z$ is assumed to be Euclidean. Latent points are
  denoted $\z_n \in \Z$, while corresponding observations are $\x_n \in \X$.
  The mapping $f: \Z \rightarrow \X$ embeds $\Z$ in $\X$; we denote
  $\M = f(\Z)$ and assume that $\M$ is a Riemannian manifold. 
  
\section{Riemannian manifolds}\label{sec:primer}
  A $d$-dimensional manifold $\M$ embedded in $\R^D (d \leq D)$ is a topological 
  space in which there exist a neighborhood around each point $\x \in \M$ that
  is homeomorphic to $\R^d$ \cite{gallot1990riemannian}. Informally, $\M$ is a (usually nonlinear) surface
  in $\R^D$ that is locally Euclidean, \ie it does not self-intersect or otherwise
  locally change dimensionality, \etc. We assume that we have a $d$-dimensional
  parametrization 
  $\Z$ of the manifold along with a mapping $f: \Z \rightarrow \X$, such that
  $\M = f(\Z)$.
  
  We first define
  the inner product between points in $\R^D$
  as $\inner{\x}{\x'} = \sfrac{1}{D}\sum_i x_i x_i'$. The division by $D$ ensures
  that the induced norm remain finite in the limit
  $D \rightarrow \infty$. Now, let $\z$ be a $d$-dimensional latent point and let
  $\Delta_1$ and $\Delta_2$ be infinitesimals, then we can compute
  their inner product around $\z$ in the data space using Taylor's Theorem,
  \begin{align}
    \inner{f(\z \! &+ \! \Delta_1) \! - \! f(\z)}{f(\z \! + \! \Delta_2) \! - \! f(\z)} \\
      &= \inner{f(\z) \! + \! \J_{\z} \Delta_1 \! - \! f(\z)}{f(\z) \! + \! \J_{\z}\Delta_2 \! - f(\z)} \\
      &= \inner{\J_{\z} \Delta_1}{\J_{\z}\Delta_2} = \sfrac{1}{D} \cdot \Delta_1\T \left( \xtx{\J_{\z}} \right) \Delta_2,
  \end{align}
  where $\J_{\z} = \partial_{\z}f \in \R^{D \times d}$ is the Jacobian of $f$ at $\z$. 
  The $d \times d$ symmetric positive definite matrix $\sfrac{1}{D}\cdot \xtx{\J_{\z}}$, thus defines
  a local inner product. We denote this matrix
  \begin{align}
    \Mz = \sfrac{1}{D}\cdot \xtx{\J_{\z}},
  \end{align}
  and refer to it as the \emph{(pull-back) metric}
  of the manifold. Note that this local inner product is invariant to reparametrizations
  of the manifold as it merely correspond to the inner product of $\X$ measured
  locally on the manifold. Hence it avoids the parametrization
  issue discussed in the opening section.

  \textbf{Distances \& interpolants.} 
  The length of a smooth curve in latent space $\vec{c}: [a, b] \rightarrow \Z$
  under the local inner product is
  \begin{align}
    \mathcal{L}(\vec{c})
       = \int_a^b \sqrt{\dot{\vec{c}}_t\T \mat{M}_{\vec{c}_t} \dot{\vec{c}}_t} \dif{t},
    \label{eq:length}
  \end{align}
  where $\dot{\vec{c}}_t = \partial_t \vec{c}(t)$ is the derivative of the curve.
  Natural interpolants (geodesics) can then be defined as length minimizing curves connecting
  two points. The length of such a curve is a natural distance measure along the manifold.
  Unfortunately, minimizing curve length gives rise to a poorly determined
  optimization problem as the length of a curve is invariant to its parametrization.
  The following proposition provides remedy \cite{gallot1990riemannian}:
  \begin{proposition}\label{prop:energy}
    Let $\vec{c}: [a, b] \rightarrow \Z$ be a smooth curve that (locally) minimizes the ``curve energy''
    \begin{align}
      \mathcal{E}(\vec{c}) &= \frac{1}{2} \int_a^b \dot{\vec{c}}_t\T \mat{M}_{\vec{c}_t} \dot{\vec{c}}_t \dif{t},
      \label{eq:energy}
    \end{align}
    then $\vec{c}$ has constant velocity and is 
    length-minimizing.
  \end{proposition}
  This energy functional is locally uniformly convex and therefore its solution
  is locally unique.
  Standard calculus of variation shows that curves of minimal energy satisfy
  the following system of second order differential equations,\hspace*{\fill}
  \begin{wrapfigure}[12]{r}{0.25\textwidth}
    \vspace{-2mm}
    \includegraphics[width=0.24\textwidth]{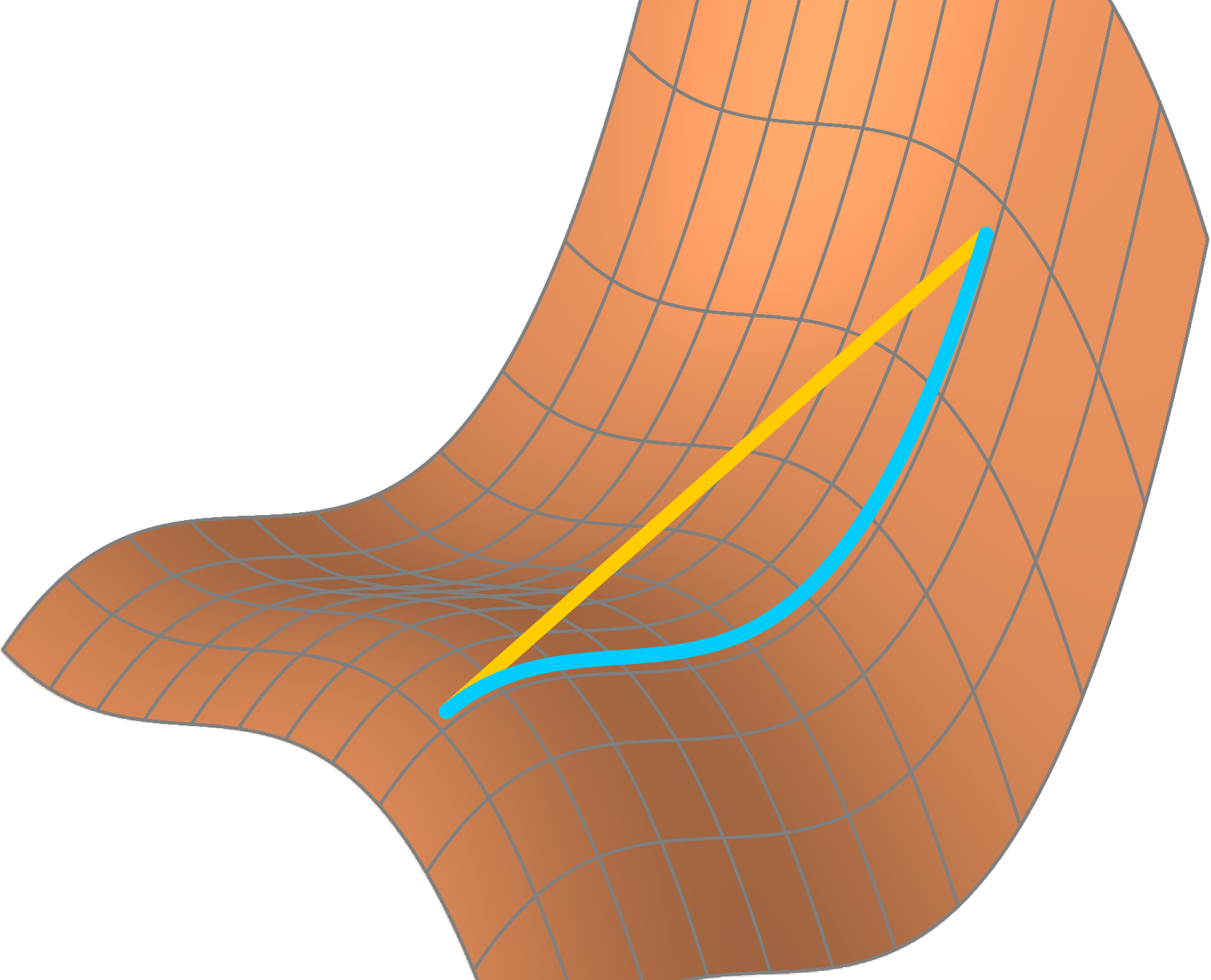}
    \vspace{-2mm}
    \caption{Geodesic interpolation along the manifold (blue) versus along a straight line (yellow).}
    \label{fig:interp}
  \end{wrapfigure}
  \begin{align}
  \begin{split}
    \label{eq:ode}
    \ddot{\vec{c}}_t 
       = -\frac{1}{2}\mat{M}_{\vec{c}_t}^{-1} & \Big[2 (\mat{I} \otimes \dot{\vec{c}}_t\T) \partial_{\vec{c}_t}\vectorize{\mat{M}_{\vec{c}_t}} \dot{\vec{c}}_t \\
      &- \partial_{\vec{c}_t}\vectorize{\mat{M}_{\vec{c}_t}}\T (\dot{\vec{c}}_t \otimes \dot{\vec{c}}_t)\Big],
  \end{split}
  \end{align}
  where $\vectorize{\cdot}$ stacks the columns of a matrix into a vector and
  $\otimes$ is the Kronecker product. Such systems can be solved numerically
  using standard techniques.
  Figure~\ref{fig:interp} gives an example geodesic.

  \textbf{Integration.}
  Given a function $h: \X \rightarrow \R$ we can integrate it over a part of the
  manifold $f(\Omega),\, \Omega \subseteq \Z$ as \cite{Pennec:JMIV:06}
  \begin{align}
    \int_{f(\Omega)} h(\x) \dif{\x}
      &= \int_{\Omega} h(f(\z)) \sqrt{\det(\mat{M}_{\z})} \dif{\z}.
  \end{align}
  The quantity $\sqrt{\det(\mat{M})}$ is known as the \emph{Riemannian volume
  measure} and is akin to the Jacobian-determinant in the
  \emph{change of variables theorem}. As before, this integration is invariant
  to reparametrizations of $\Z$ as it is performed with respect to the measure of $\X$.
  
\section{Manifold learning with kernels}\label{sec:kernels}
  We now consider data $\x_{1:N}$ distributed on a compact $d$-dimensional Riemannian
  submanifold $\M \subset \R^D$ embedded in the data space. We consider a
  known set of $d$-dimensional representations $\z_{1:N}$ and estimate the
  mapping $f: \Z \rightarrow \X$ using kernel methods.
  Note that this
  manifold is only locally diffeomorphic to $d$-dimensional Euclidean space,
  and it may globally self-intersect\footnote{Technically, this render the manifold
    \emph{immersed} rather than \emph{embedded}; the distinction is not important
    for our purposes.}.
  For the sake
  of analysis, we assume noise-free data and consider the limit $N \rightarrow \infty$.
  This setting is sufficient to prove our main point, but the analysis also hold under noise.
  
  Our key question is if we can recover the true Riemannian structure of $\Z$.
  Methods that fail at this given infinite noise-free data
  should be avoided. To give an answer, we will study the metric
  in regions that are near the training data, and in regions that are far away.
  We formalize this as follows.
  \begin{definition}
    For a point $\z$ and a dataset $\mat{Z}$, the distance between them is
    $  \dist(\z, \mat{Z}) = \inf_{\tilde{\z} \in \mat{Z}} \| \z - \tilde{\z} \| $.
    Note that this infimum always exist as the element-wise distance is bounded from below by 0.
  \end{definition}
  %
  %
  \begin{definition}
    For a function $\x = h(\z)$, we define the limits
    \begin{align}
      \x \away \hat{\x}
      \quad \mathrm{and} \quad
      \x \near \hat{\x}
    \end{align}
    if for any sequences $\hat{\z}_l$ such that
    $\dist(\hat{\z}_l, \mat{Z}) \xlongrightarrow{l \rightarrow \infty} \infty$,
    or
    $\dist(\hat{\z}_l, \mat{Z}) \xlongrightarrow{l \rightarrow \infty} 0$,
    respectively,    
    we have $h(\hat{\z}_l) \xlongrightarrow{l \rightarrow \infty} \hat{\x}$. 
    Note that these limits are not always defined.
  \end{definition}

\begin{figure*}
  \includegraphics[width=0.24\textwidth]{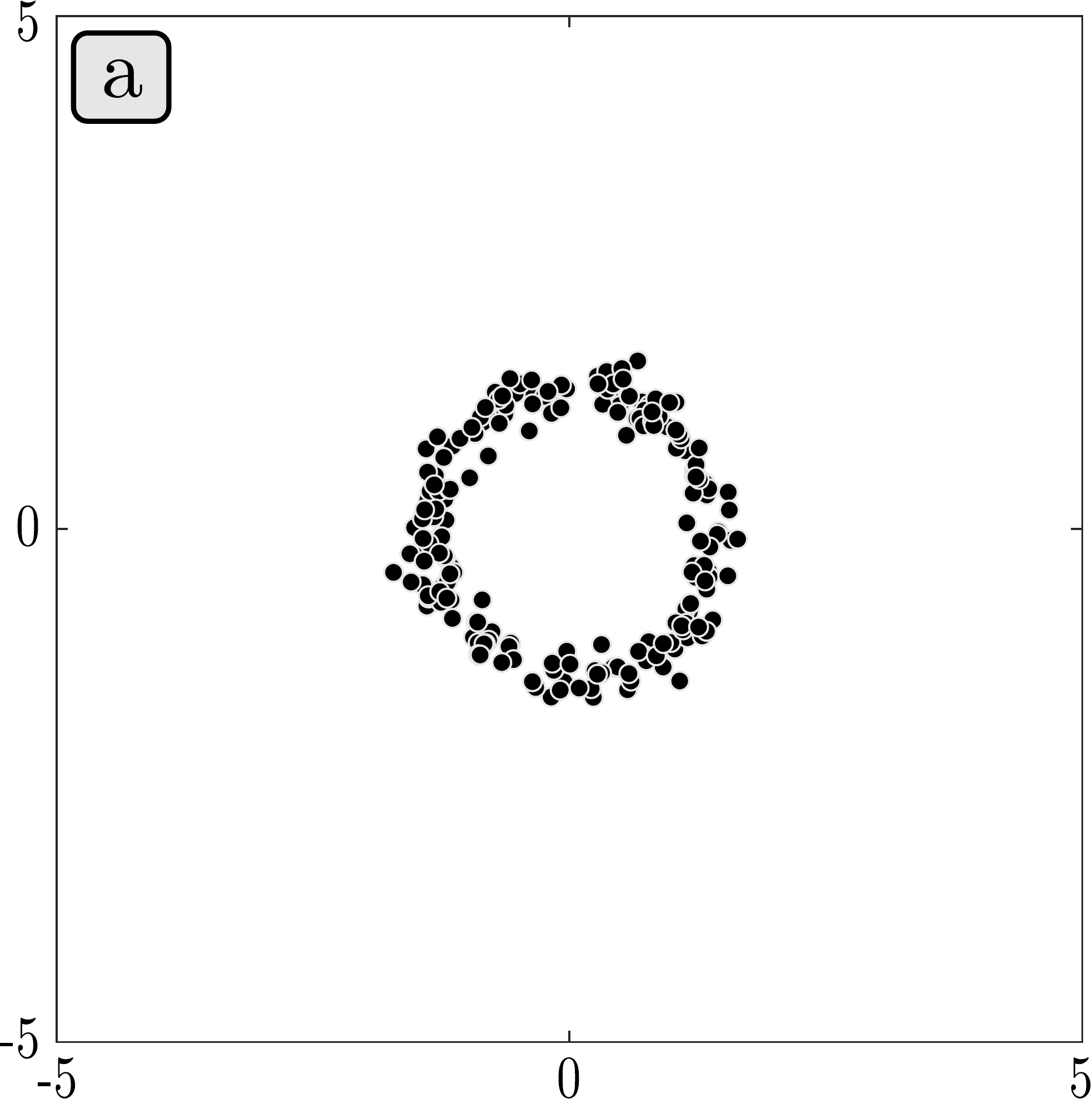}
  \includegraphics[width=0.24\textwidth]{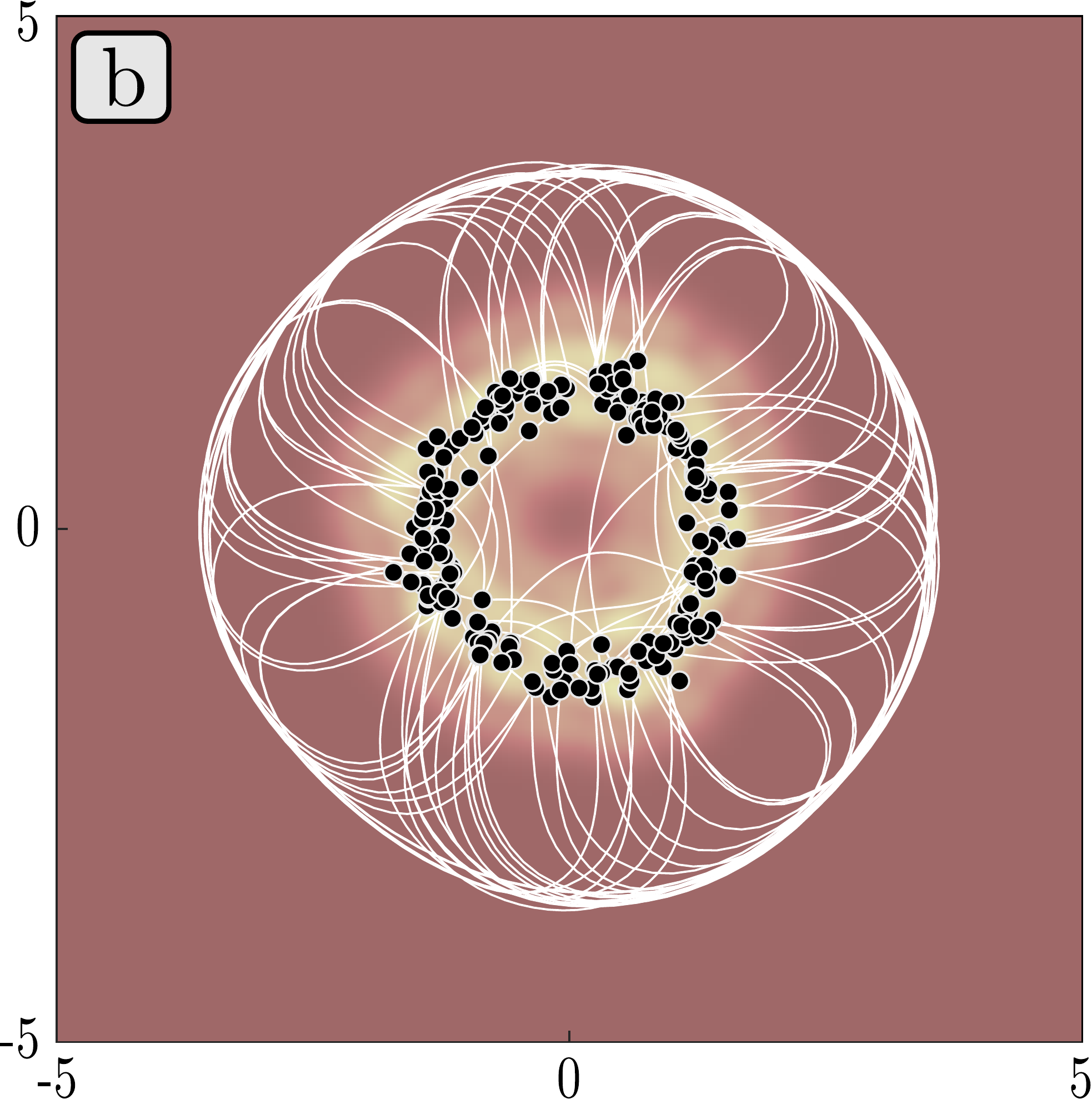}
  \includegraphics[width=0.24\textwidth]{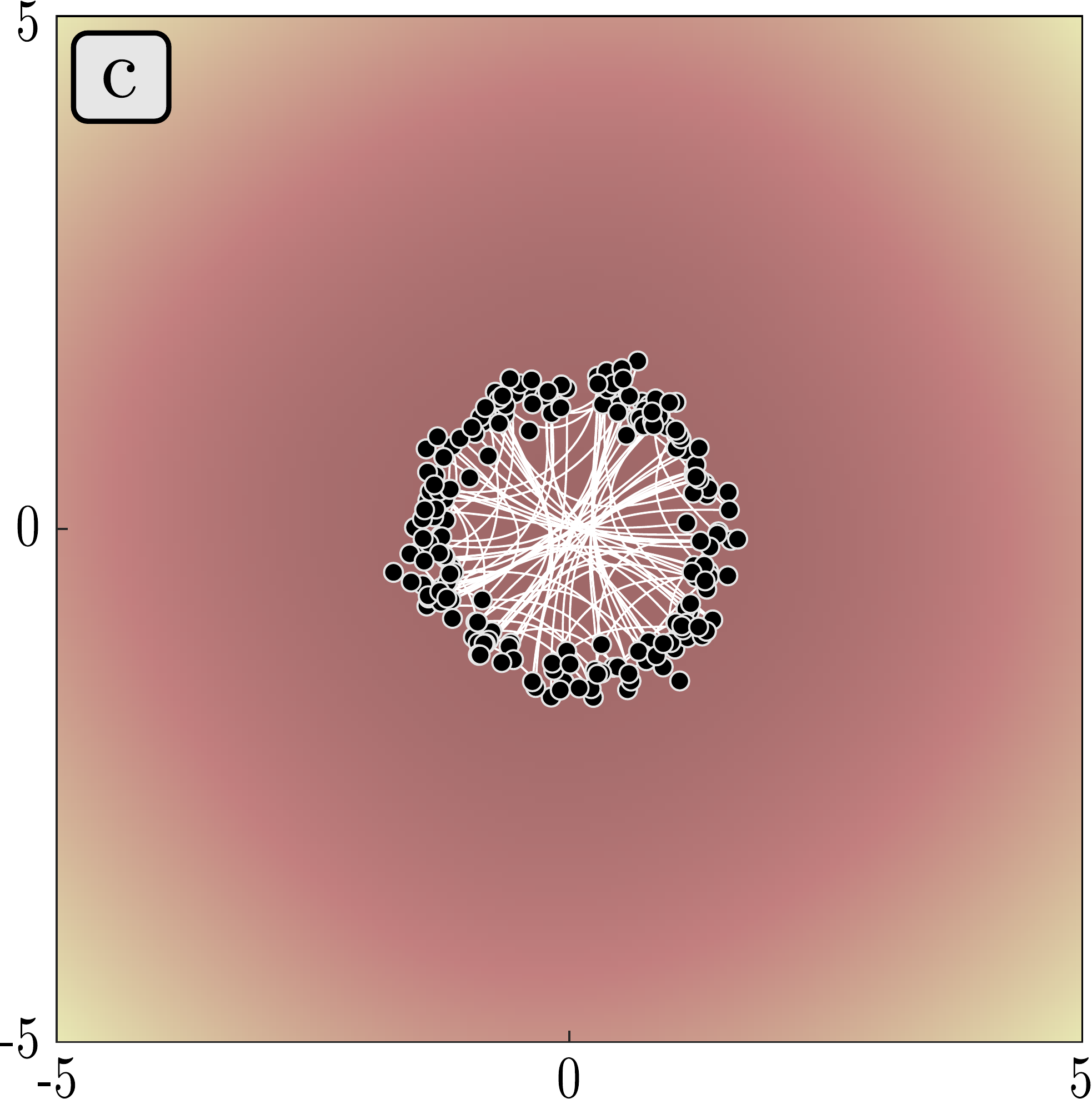}
  \includegraphics[width=0.24\textwidth]{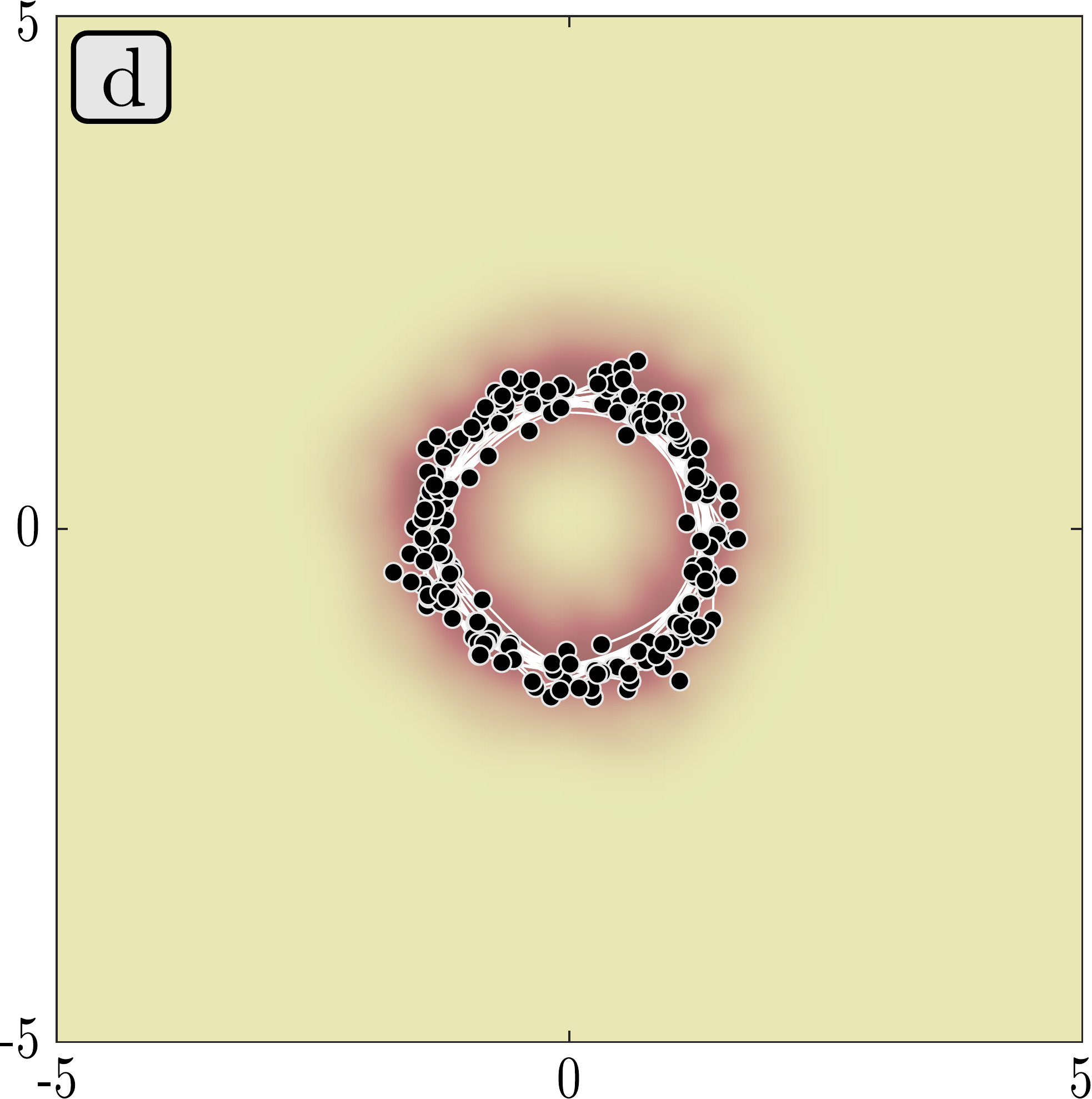}
  \caption{(a) Latent points $\z_n \in \Z$ for our guiding example.
           (b) Geodesics for Gaussian kernel ridge regression. These are pushed away
               from the data via ``teleports''.
           (c) Geodesics for kernel ridge regression with a Gaussian+linear kernel.
               The linear extrapolation implies (almost) linear geodesics.
           (d) Geodesics for Gaussian process regression. The uncertainty make
               geodesics move along the manifold.}
  \label{fig:big1}
\end{figure*}
\paragraph{A guiding example.}
  To illustrate our main point, we draw data uniformly on a unit circle and nonlinearly embed it in 
  $\X = \R^{1000}$ with added Gaussian noise. We project this data into $\Z = \R^2$ while
  keeping the circular structure of data and
  learn a mapping $f$ from $\Z$ to $\X$. Details of this process are in Appendix~\ref{app:data}.
  Finally, we compute shortest paths
  under the pull-back metric; if the true metric is recovered we should see shortest
  paths corresponding to circular arcs in $\Z$. 
  Figure~\ref{fig:big1}a show the latent points in $\Z$.
  
\subsection{The deterministic setting}\label{sec:deterministic}
  We now consider learning the mapping $f: \Z \rightarrow \X$
  using \emph{kernel ridge regression} \cite{learning_with_kernels}, \ie
  \begin{align}
    f_{\text{krr}}(\z_*)
      &= k_{*, \z} \left(k_{\z, \z} + \sigma^2 \mat{I}\right)^{-1} \mat{X}, 
  \end{align}
  where $k$ is a suitable kernel function, $\mat{X} \in \R^{N \times D}$ is the data matrix, and we have used the short-hand notations
  $k_{*, \z} = k(\z_*, \z_{1:N}) \in \R^{1 \times N}$ and
  $k_{\z, \z} = k(\z_{1:N}, \z_{1:N}) \in \R^{N \times N}$. 
  Since we consider noise-free data, we have $\sigma^2 = 0$.
  The pull-back metric associated with this regression function is 
  \begin{align}
    \mat{M}_{\text{krr}}(\z_*)
      &= \partial_{\z_*}k_{*, \z} k_{\z, \z}^{-1} \mat{X} \mat{X}\T k_{\z, \z}^{-1} \partial_{\z_*}k_{*, \z}\T.
  \end{align}
  Assuming a universal kernel \cite{learning_with_kernels}, then $f_{\text{krr}}$
  will correspond to the true mapping where we have data when $N \rightarrow \infty$.
  Consequently, we recover the true metric where we have data, \ie
  $\mat{M}_{\text{krr}} \near \mat{M}_{\text{true}}$.
  

  \paragraph{Teleports?}
  The behavior away from data depend on the kernel. We first consider
  the Gaussian kernel
  \begin{align}
    k_{\text{RBF}}(\z, \z') = \theta_{\text{RBF}} \cdot \exp\left( -\frac{\alpha}{2}\|\z - \z'\|^2 \right),
    \label{eq:rbf_kernel}
  \end{align}
  but note that similar observations hold for most common stationary kernels.
  From this, we see that
  \begin{align}
    f_{\text{RBF}} \away \vec{0}
    \qquad\text{and}\qquad
    \mat{M}_{\text{RBF}} \away \mat{0}.
    \label{eq:rbf_metric}
  \end{align}

  To illustrate the geometric implication of this observation, we consider our
  guiding example. We compute shortest paths
  under the pull-back metric; if the true metric is recovered these should be
  circular arcs in $\Z$. Figure~\ref{fig:big1}b
  show the recovered geodesics; 
  we see that they systematically shy away
  from the data and generally do not resemble circular arcs. The explanation is
  simply that in terms of length-minimization, it is ``free'' to move through
  regions where the metric is zero. The result in Eq.~\ref{eq:rbf_metric}, thus, 
  implies that geodesics are encouraged to move away from the data. 
  Intuitively, we can think of regions in $\Z$ without data as ``teleports''
  that points can move freely between.
  This also hold true, when the manifold is densely sampled, and consequently
  geodesics will generally not move along the data manifold: \emph{the manifold
  geometry is not recovered}.
  
  \paragraph{Flat manifolds?}
  These teleports appear because the chosen kernel cause $f$
  to extrapolate to a constant. We now consider a kernel that extrapolate linearly,
  \begin{align}
    k_{\text{RBF+lin}}(\z, \z')
      &= k_{\text{RBF}}(\z, \z') + \theta_{\text{lin}} \z\T \z'.
  \end{align}
  Similarly to before, we see that
  \begin{align}
    f_{\text{RBF+lin}} \away \theta_{\text{lin}} \z_*\T\mat{Z} k_{\z, \z}^{-1} \mat{X} = \z_*\T \mat{B},
  \end{align}
  where $\mat{B} = \theta_{\text{lin}} \mat{Z} k_{\z, \z}^{-1} \mat{X} \in \R^{d \times D}$.
  This amounts to linear extrapolation, as expected. When we move away from the data, the
  metric then becomes
  \begin{align}
    \mat{M}_{\text{RBF+lin}} \away \mat{B}\mat{B}\T.
  \end{align}
  This is a (scaled) Euclidean metric, implying that the learned manifold is
  flat in regions where we do not have data. As the pull-back metric measure
  distances in $\X$, where straight lines are shortest curves, then geodesics
  on the learned manifold will be encouraged to go through the flat regions
  where data is missing. This is also evident in Fig.~\ref{fig:big1}c that
  shows the results of our guiding example. Here we see that geodesics are almost
  straight lines, implying that the learned manifold did not recover the
  structure of the data.
  
  \paragraph{A regularization perspective.}
  The phenomenon can also be understood by considering regression functions $f$ that
  minimize a local (log) likelihood function \citep{localregression, local-tibshirani}
  \begin{align}
    \mathcal{L} &= \sum_{n=1}^N w_n \mathcal{L}_n + \lambda\phi[f],
    \label{eq:local_loglike}
  \end{align}
  where $\mathcal{L}_n$
  is the loss associated with the $n^{\text{th}}$ observation, $w_n$ is a weight that
  decays with the distance to the point where $f$ is evaluated, and $\phi$ is the
  regularizer. \citet{girosi1995regularization} show
  that most regularizers are low-pass filters (thereby avoiding
  ``wiggly'' regression functions). Away from the data, the regularizer dominate
  the above cost function, implying a strongly low-pass filtered regression function.
  By Parseval's theorem \cite{stein2000digital}
  this reduce the energy of curves that move away from the data, which bias
  geodesics to move away from the data.
  
  \paragraph{An unstable solution?}
  We have seen that the traditional constant and linear extrapolation
  schemes imply that we cannot learn the correct geometry: either we introduce
  teleports or we learn mostly flat manifolds. We now 
  ask: \emph{how should we extrapolate in order to learn the correct geometry?}
  For geodesics to stay on the manifold, inner products must take large values away from the data.
  \begin{align}
    \mat{M} \away \text{[sufficiently large]}.
  \end{align}
  That is, to ensure that geodesics always stay on the manifold, length-minimization
  must be penalized sufficiently for leaving the manifold. We, currently, do not
  have a tight bound on how large the metric must be to ensure this, though a loose
  bound is provided by the radius of the manifold. That is, let
  \begin{align}
    r &= \sup_{\z, \z' \in \M} \dist(\z, \z')
  \end{align}
  denote the largest distance between points on the manifold, then geodesics stay on the manifold if 
  \begin{align}
    \lambda_{\text{min}}\left(\mat{M}_{\text{sufficient}}\right) \away r^2.
  \end{align}
  Here $\lambda_{\text{min}}$ denote
  the function returning the smallest eigenvalue of a matrix.
  In differential geometry, it is common
  to call \emph{any} locally length-minimizing curve a geodesic. Here we mean
  the shortest geodesic. To ensure that \emph{all} locally length-minimizing curves
  stay on the manifold, we have no tighter bound than
  \begin{align}
    \lambda_{\text{min}}\left(\mat{M}_{\text{ideal}}\right) \away \infty.
  \end{align}
  If the metric must extrapolate to a large matrix, then the Jacobian $\J$
  must also extrapolate to a large matrix (since $\mat{M} \propto \xtx{\J}$).
  The results of \citet{girosi1995regularization} dictate that regularizing
  towards functions with large derivatives imply that the solution is no longer
  low-pass filtered (\ie it will ``wiggle'').
  Regularizing towards large derivatives, thus, go against common wisdom as regression functions with large derivatives
  in regions of little data, will generally not be stable.

  \paragraph{Summarizing discussion.}
  Most common choices for estimating $f$ will ensure that the geometry of $\M$
  is well-estimated near the data, so the key factor to determine if we can
  well-estimate the manifold geometry is the behavior of $f$ away from the data.
  In these regions, we depend on prior assumptions on $f$ to determine the geometry.
  The most common priors are related to the
  smoothness of $f$, where we use the terminology of \citet{girosi1995regularization},
  \ie a function is considered ``smooth'' if its spectrum is dominated by low-frequency
  components. We have seen that the smoother assumptions we
  are willing to make, the more geodesics are drawn away from the data (``off the manifold'', so to say).
  
  \begin{wrapfigure}[12]{r}{0.25\textwidth}
    \vspace{-2mm}
    \includegraphics[width=0.24\textwidth]{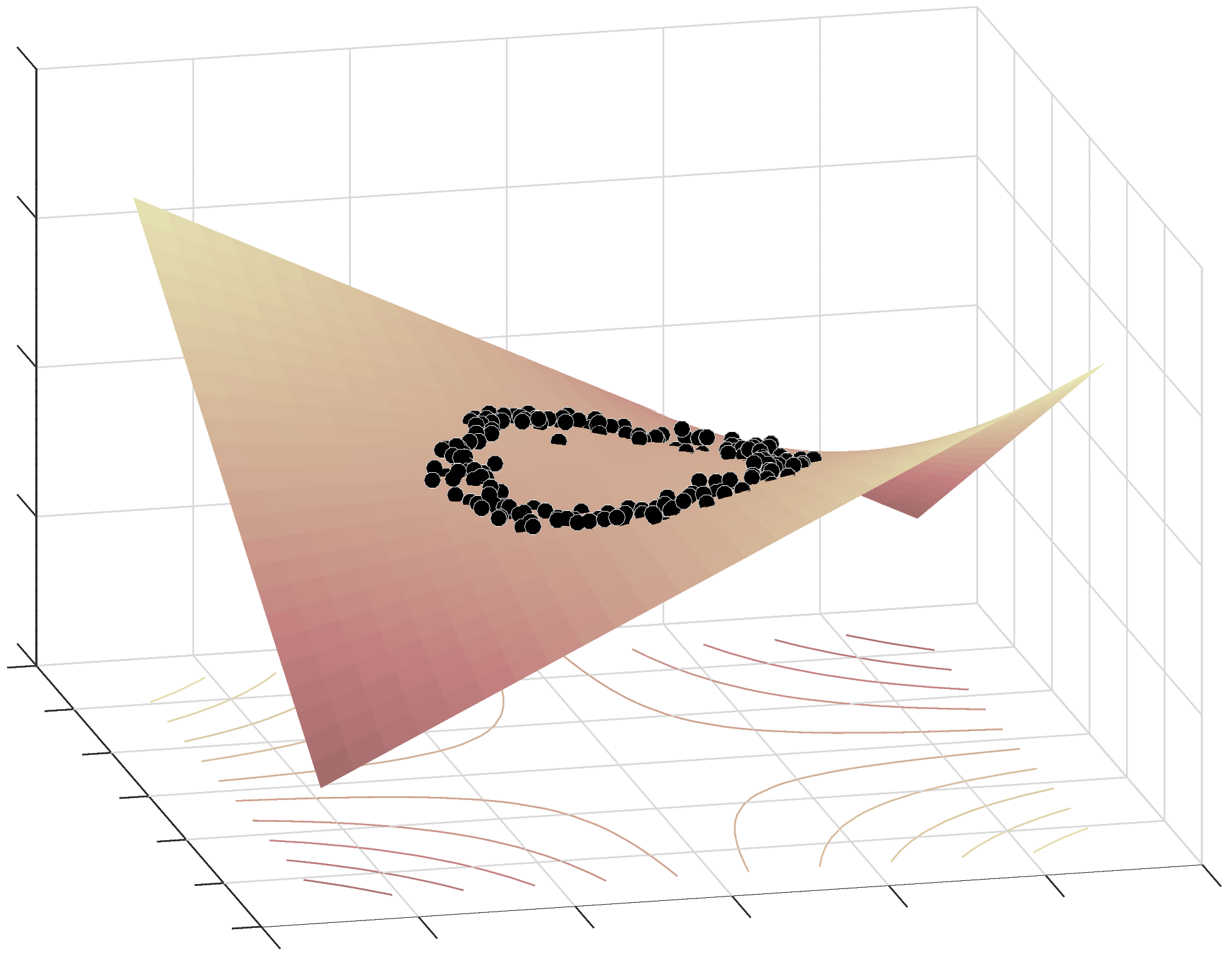}
    \vspace{-2mm}
    \caption{Smooth interpolation of data imply that shortest paths measured in data space
      cut across holes.}
    \label{fig:smoothness}
  \end{wrapfigure}
  As a specific example, consider a local likelihood \eqref{eq:local_loglike} with
  regularizer
  \begin{align}
    \phi[f]
      = \E\left[ \left\| \frac{\partial f}{\partial \z} \right\|^2 \right]
      = \E\left[ \trace\xtx{\J} \right]
      = D \E\left[ \trace\mat{M} \right] .
  \end{align}
  \citet{bishop1995training} has shown that this is (approximately) the regularizer
  implied by training under additive noise.\footnote{As an example, the mean decoder of a VAE is trained under this regularizer.}
  We see that this common regularizer
  imply a ``small'' metric, when moving away from the data, which in turn imply
  short geodesic segments away from the data. Minimization of \emph{curve energy}
  \eqref{eq:energy} --- or equivalently \emph{curve length} \eqref{eq:length} ---
  will, thus, be biased towards regions of no data as this is where the metric
  is minimal.
  
  The general phenomenon is easily understood by considering a data manifold with a hole.
  If the applied regression function is very smooth, then the hole will be
  interpolated almost linearly, which imply that shortest paths along the manifold
  will cross over the hole.
  In the end, we are, thus, left with a simple choice: \emph{either give up on learning the manifold geometry correctly}
  (by assuming $f$ is very smooth) or \emph{give up on stable learning} (by assuming
  $f$ is not very smooth). Neither choice is desirable.

%
%

\subsection{The Bayesian setting}\label{sec:probabilistic}
  As a  natural probabilistic extension of the previous sections we now let
  $f$ consist of component-wise conditionally
  independent Gaussian processes (GPs) \cite{rasmussen:book},
  \begin{align}
    f_i(\z) &\sim \mathcal{GP}(m_i(\z), k(\z, \z')),
    \quad \forall i = 1, \ldots, D.
  \end{align}
  This is a \emph{Gaussian Process latent variable model (GP-LVM)} \cite{gplvm}.
  Here $m_i$ and $k$ are the mean and covariance functions of the $i^{\mathrm{th}}$
  GP. Note that the posterior mean function coincide with the previously
  considered kernel ridge regression model.
  Like \citet{gplvm}, we assume the same covariance function across all dimensions
  to simplify calculations. The geometry of this model
  was first studied by \citet{Tosi:UAI:2014}. 

  The pull-back metric $\mathbf{M} = \sfrac{1}{D} \xtx{\J}$ is now
  a stochastic Riemannian metric since $f$ is stochastic.
  As Gaussian variables are closed under differentiation, then $\mathbf{J}$
  is Gaussian, $\J \sim \prod_{j=1}^D \N (\mu(j,:) , \bs{\Sigma})$, and
  $\mat{M}$ follows a non-central Wishart distribution \cite{Tosi:UAI:2014, muirhead} 
  \begin{align}
    D\cdot\mat{M} \sim
    \mathcal{W}_d(D, \bs{\Sigma}, \bs{\Sigma}^{-1}\E[\J]\T \E[\J]).
  \end{align}
  The entire metric by definition follows a generalized Wishart process \cite{wilson2010generalised, hauberg:owr:2018}.
  A sample path from this process is smooth when the covariance
  function $k$ is also smooth, and we have a proper distribution over Riemannian
  metrics. However, a sample from $f$ gives a manifold that is only locally
  diffeomorphic to $d$-dimensional Euclidean space, and it may globally self-intersect. 

  Since the metric is a stochastic variable, we cannot apply standard Riemannian geometry
  to understand the space $\Z$ (curvature is stochastic,
  geodesics are stochastic, etc).
  We can, however, inspect the leading moments of the metric
  \begin{align}
    \E[\mat{M}] 
      &= \frac{1}{D} \E[\xtx{\J}]
       = \frac{1}{D} \E[\J]\T \E[\J] + \bs{\Sigma} \label{eq:exp_metric} \\
    \var{M_{ij}} \!
      &= \! \frac{\Sigma_{ij}^2 + \Sigma_{ii}\Sigma_{jj}}{D}
      \! + \! \frac{\mu_j\T \bs{\Sigma} \mu_j}{D^2}
      \! + \! \frac{\mu_i\T \bs{\Sigma} \mu_i}{D^2} .
  \end{align}
  We see that $\E[\mat{M}]$ remain strictly positive, while
  $\var{M_{ij}} = \mathcal{O}\left(\sfrac{1}{D}\right)$ vanishes in the limit $D \rightarrow \infty$.
  This can equivalently be seen from the central limit theorem.
  In high dimensions, the metric, thus, becomes deterministic even if the underlying manifold is
  stochastic. This is useful as it implies that we can well-approximate
  the stochastic metric with a well-understood deterministic metric.
  
  To see if this approach can learn the geometric structure of the data manifold,
  we again consider the Gaussian kernel \eqref{eq:rbf_kernel}. Straight-forward
  calculations show that
  \begin{align}
    \bs{\Sigma} \near \mat{0}
    \qquad\text{and}\qquad
    \bs{\Sigma} 
       \away \alpha\theta_{\text{RBF}}\mat{I},
  \end{align}
  where $\alpha$ and $\theta_{\text{RBF}}$ are the kernel parameters.
  From this we see that near the data, the expected metric \eqref{eq:exp_metric}
  coincides with the true pull-back metric of the manifold (as in the deterministic
  setting),
  \begin{align}
    \hspace{-15mm}
    \mathbb{E}[\mathbf{M}] \near \frac{1}{D} \mathbb{E}[\mathbf{J}]\T \mathbb{E}[\mathbf{J}].
  \end{align}  
  In regions of $\Z$  where there is no data, we have
  \begin{align}
    \mathbb{E}[\mathbf{M}] \away \frac{1}{D} \mathbb{E}[\mathbf{J}]\T \mathbb{E}[\mathbf{J}] + \alpha\theta_{\text{RBF}}\mat{I}.
  \end{align}
  If $\alpha\theta_{\text{RBF}}$ is sufficiently large then geodesics will
  not go through regions of $\Z$ where we do not have data. When data is sampled
  densely on the manifold, we often estimate large values of $\alpha$
  (corresponding to a less smooth manifold), and a
  large penalty will be payed for ``falling off the manifold''.
  To validate these observations,
  we return to our guiding example; Fig.~\ref{fig:big1}d shows that
  geodesics under the expected metric \eqref{eq:exp_metric} actually follow
  approximately circular arcs.
  This aligns with the theoretical analysis and demonstrates that, unlike
  a deterministic method, a probabilistic method can actually learn the
  differential geometric structure of a data manifold.
  This is, however, no guarantee: we can only accurately learn
  the manifold geometry when data is sampled sufficiently dense on the manifold,
  but now there is hope, whereas deterministic approaches are bound to fail.
  
  \paragraph{Discussion.}
  Deterministic methods can capture local geometry of the data
  manifold near the observed data, but they fail to capture the geometry
  where data is missing. This is not surprising, as we can generally
  only learn when we have data. What is, perhaps, more surprising is that
  if we can estimate the uncertainty of the manifold, then that translate
  directly into geometric information: if there is a hole in the manifold,
  then we can only see it through a lens of uncertainty. Not quantifying the
  uncertainty prevents us from seeing holes and boundaries of a data
  manifold. \emph{Uncertainty, thus, plays the same role as topology} in classic
  geometry, and this must also be estimated from data.

\section{Bayesian geometry}\label{sec:bayes_geom}
  As the expected metric can capture the geometry
  of the data manifold, we seek a better understanding of 
  stochastic Riemannian metrics.
  We here consider the case of the GP-LVM where $f$ is a smooth GP. 
  
  \subsection{Detour: Euclidean geometry}
  Before analyzing stochastic Riemannian metrics, 
  consider a stochastic Euclidean metric. Let
  $\vec{u}, \vec{v} \in \R^d$ denote deterministic vectors, and define their
  stochastic inner product 
  \begin{align}
    \inner{\vec{u}}{\vec{v}}
      &= \big(\mat{A}\vec{u}\big)\T\big(\mat{A}\vec{v}\big)
       = \vec{u} \big(\xtx{\mat{A}}\big) \vec{v},
  \end{align}
  where $\mat{A} \in \R^{D \times d}$ is a matrix-valued random variable. 
%
  Under this inner product, the shortest path between two
  points is the straight line, so stochasticity does not 
  change our usual intuitions. The length of this line is,
  however, stochastic. Its expectation is found by letting
  $\bs{\Delta} = \mat{A}(\vec{u} - \vec{v})$, then
  \begin{align}
    \E\!\left[\|\bs{\Delta}\|^2\right]
      &= \E[\bs{\Delta}]\T \E[\bs{\Delta}]
       + \trace\left( \cov{\bs{\Delta}} \right).
  \end{align}
  The expected distance under a stochastic Euclidean
  metric grows with both mean and variance of the basis $\mat{A}$.
  Hence, expected distances are inherently large when the basis (or equivalently the metric)
  has large variance.
  
  \begin{exbox}
  \begin{example}[a Gaussian basis]\label{ex:gaussian_basis}
  Let each row of $\mat{A}$ be drawn independently from
  $\N(\vec{0}, \bs{\Sigma})$, such that the metric
  follows a Wishart distribution \cite{muirhead},
  \begin{align}
   \mat{M}
      = \xtx{\mat{A}}
       \sim \mathcal{W}_d(D, \bs{\Sigma}).
  \end{align}
  Then the distance from $\vec{u}$ to $\vec{v}$ is Nakagami distributed \cite{laurenson1994nakagami}
  \begin{align}
    \dist(\vec{u}, \vec{v})
      &\sim \text{Nakagami}\left( \frac{D}{2}, D\sigma_{\vec{u}, \vec{v}}^2 \right),
  \end{align}
  and the expected distance is \cite{hauberg:nakagami:2018}
  \begin{align}
    \E\left[ \dist(\vec{u}, \vec{v}) \right]
      &= \frac{\Gamma\left( \frac{D+1}{2} \right)}{\Gamma\left( \frac{D}{2} \right)}
         \sqrt{2} \sigma_{\vec{u}, \vec{v}}
     \propto \sigma_{\vec{u}, \vec{v}}, \\
    \text{where}\quad
    \sigma_{\vec{u}, \vec{v}}^2
      &= (\vec{u}-\vec{v})\T \bs{\Sigma} (\vec{u}-\vec{v}).
  \end{align}
  We see that the expected distance correspond to
  Mahalanobis' distance using the inverse covariance. Also note that scaling
  $\bs{\Sigma}$ also scales the expected distance: \emph{very uncertain
  metrics imply large expected distances.}
  \end{example}
  \end{exbox}

  \subsection{Geodesics}
  We now return to the geometry of the GP-LVM, and seek to understand shortest paths.
  Let $\vec{c}: [a, b] \rightarrow \Z$ denote a deterministic differentiable
  curve, and let $f(\vec{c})$ denote its stochastic embedding in $\X$.
  We stress that $\vec{c}$ is a deterministic curve in $\Z$, while $f(\vec{c})$
  is a GP in $\X$.
  %
  The energy \eqref{eq:energy} of $f(\vec{c})$ is a random quantity and it is natural to consider its expectation with
  respect to the random metric.
  Since the energy integrand is positive, Tonelli's Theorem tells us that this expected energy is
  \begin{align}
    \bar{\mathcal{E}}(\vec{c})
      &\equiv \E_{\mat{M}}\left[\mathcal{E}(f(\vec{c}))\right]
       = \frac{1}{2} \E_{\mat{M}}\left[\int_a^b \dot{\vec{c}}_t\T \mat{M}_{\vec{c}_t} \dot{\vec{c}}_t \dif{t}\right] \\
      &= \frac{1}{2} \int_a^b \dot{\vec{c}}_t\T \E\left[\mat{M}_{\vec{c}_t}\right] \dot{\vec{c}}_t \dif{t}.
  \end{align}
  This implies that the curve $\vec{c}$ with minimal expected energy over the stochastic
  manifold, is the geodesic
  under the deterministic Riemannian metric $\E\left[ \mat{M} \right]$.
  This is exactly the metric considered in Sec.~\ref{sec:probabilistic}.
  
  We can understand the curve minimizing expected energy in more explicit terms
  as follows. Let $u_t = \E[\| \dot{\vec{c}}_t \|]$ and $v_t = 1$ denote two functions
  over the interval $[a, b]$; here we use the short-hand notation
  $\| \dot{\vec{c}}_t \| = \sqrt{\dot{\vec{c}}_t\T \mat{M}_{\vec{c}_t} \dot{\vec{c}}_t}$.
  Then Cauchy-Scwartz's inequality tells us that
  \begin{align}
    \left| \inner{u}{v} \right|^2
      &\leq \|u\|^2 \cdot \|v\|^2 \\
    \left( \int_a^b \E[\| \dot{\vec{c}}_t \|] \dif{t} \right)^2
      &\leq  \int_a^b \E[\| \dot{\vec{c}}_t \|]^2 \dif{t} \cdot \int_a^b \dif{t} \\
      &= (b-a) \int_a^b \E[\| \dot{\vec{c}}_t \|]^2 \dif{t}.
  \end{align}
  Let
  \begin{align}
    \bar{\mathcal{L}}(\vec{c})
      &= \E \left[ \int_a^b \| \dot{\vec{c}}_t \| \dif{t} \right]
       = \int_a^b \E[\| \dot{\vec{c}}_t \|]\dif{t}
  \end{align}
  denote the expected length of $\vec{c}$ then
  \begin{align}
    \int_a^b \E[\| \dot{\vec{c}}_t \|]^2 \dif{t}
      &\geq 
    \frac{\bar{\mathcal{L}}^2(\vec{c})}{b-a}.
  \end{align}
  Equality is achieved when $u_t$ and $v_t$ are parallel, that is when
  $\E[\| \dot{\vec{c}}_t \|]$ is constant. We can always reparametrize $\vec{c}_t$
  to have constant expected speed and achieve equality.
  Since $\var{x} = \E[x^2] - \E[x]^2$, we see that
  \begin{align}
    \nonumber
    \int_a^b \!\E[\| \dot{\vec{c}}_t \|]^2 \dif{t}
      &= \int_a^b \! \E[\| \dot{\vec{c}}_t \|^2] \dif{t}
       - \int_a^b \var{\| \dot{\vec{c}}_t \|} \dif{t} \\
      &= 2\bar{\mathcal{E}}(\vec{c})
       - \int_a^b \var{\| \dot{\vec{c}}_t \|} \dif{t}.
  \end{align}
  Assuming that the curve has been parametrized to have constant expected speed,
  we then get
  \begin{align}
    \bar{\mathcal{E}}(\vec{c})
      &= 
    \frac{\bar{\mathcal{L}}^2(\vec{c})}{2(b-a)} +
    \frac{1}{2}\int_a^b \var{\| \dot{\vec{c}}_t \|} \dif{t}.
    \label{eq:energy_length_random}
  \end{align}
  Minimizing expected curve energy, thus, does not always minimize the expected
  curve length. Rather, this balances the minimization of expected curve length
  and the minimization of curve variance.
  
  \paragraph{Implications and interpretation.}
  On a deterministic manifold,  minimizing curve energy results in a
  curve of minimal length (by Proposition~\ref{prop:energy}). When the manifold is stochastic, we see that minimizing expected
  curve energy does not imply a minimization of expected curve length. In some sense,
  this is disappointing. Yet, it is intriguing that minimizing expected energy 
  corresponds to minimizing a combination of length and variance. Since the expected energy minimizing curve is
  the geodesic under the expected Riemannian metric, this also lend itself
  to easy computation.
  
  \begin{exbox}
  \begin{example}[a GP prior manifold]\label{ex:gp_prior_manifold}
  There are cases where expected energy and length are strongly related.
  Let each dimension of $f$ be a zero-mean
  GP with a sufficiently smooth covariance function. Such processes
  are common for specifying priors. Let the Jacobian of $f$ at $\z$ be 
  \begin{align}
    \mat{J}_{\z} \sim \prod_{j=1}^D \N(\vec{0}, \bs{\Sigma}_{\z}),
  \end{align}
  such that 
  \begin{align}
    D\cdot\mat{M}_{\z} = \xtx{\J_{\z}}
      \sim \mathcal{W}_d(D, \bs{\Sigma}_{\z}).
  \end{align}
  The expected energy of a curve $\vec{c}: [a, b] \rightarrow \Z$ is then
  \begin{align}
    \bar{\mathcal{E}}(\vec{c})
      &= \frac{1}{2} \int_a^b \E\left[ \dot{\vec{c}}_t\T \mat{M}_{\vec{c}_t} \dot{\vec{c}}_t \right] \dif{t}
       = \frac{D}{2} \int_a^b \dot{\vec{c}}_t\T \bs{\Sigma}_{\vec{c}_t} \dot{\vec{c}}_t \dif{t}
    \label{eq:gp_prior_energy}
  \end{align}
  since $\dot{\vec{c}}_t\T \mat{M}_{\vec{c}_t} \dot{\vec{c}}_t \sim \mathcal{W}_1(D, \dot{\vec{c}}_t\T \bs{\Sigma}_{\vec{c}_t} \dot{\vec{c}}_t)$.
  Following Example~\ref{ex:gaussian_basis} we see that the
  expected length of $\vec{c}$ is
  \begin{align}
    \bar{\mathcal{L}}(\vec{c})
      &\propto \int_a^b \sqrt{\dot{\vec{c}}_t\T \bs{\Sigma}_{\vec{c}_t} \dot{\vec{c}}_t} \dif{t}.
    \label{eq:gp_prior_len}
  \end{align}
  We can then interpret $\bs{\Sigma}_{\z}$ as a Riemannian metric and note that
  the expressions for curve length and energy under this metric 
  correspond to Eqs.~\ref{eq:gp_prior_len} and \ref{eq:gp_prior_energy}.
  By Proposition~\ref{prop:energy}, we then have that minimizing expected curve energy also minimize
  expected curve length.
  \end{example}
  \end{exbox}


    %
  \subsection{Integration}
    Let $\Omega \subseteq \Z$ such that $f(\Omega) \subseteq \M$, and let
    $h: f(\Omega) \rightarrow \R$ denote a real-valued integratable function over the specified part of the manifold.
    We now seek its integral over the entire domain under a random Riemannian metric $\mat{M}$. That is,
    \begin{align}
      \bar{h}
        &= \int_{f(\Omega)} h(\x) \dif{\x} = \int_{\Omega} h(f(\z)) \sqrt{\det\mat{M}_{\z}} \dif{\z}.
    \end{align}
    Since the metric is stochastic, then so is $\bar{h}$. We evaluate its
    expectation as
    \begin{align}
      \E_{\mat{M}}[\bar{h}]
        &= \E_{\mat{M}}\left[ \int_{\Omega} h(f(\z)) \sqrt{\det\mat{M}_{\z}} \right] \dif{\z} \\
        &= \int_{\Omega} h(f(\z)) \E\left[ \sqrt{\det\mat{M}_{\z}} \right] \dif{\z}.
     \label{eq:exp_integral}
    \end{align}
    That is, the integration is simply performed under the expected volume measure
    (as expected).
    
    \begin{wrapfigure}[14]{r}{0.25\textwidth}
      \vspace{-4mm}
      \includegraphics[width=0.24\textwidth]{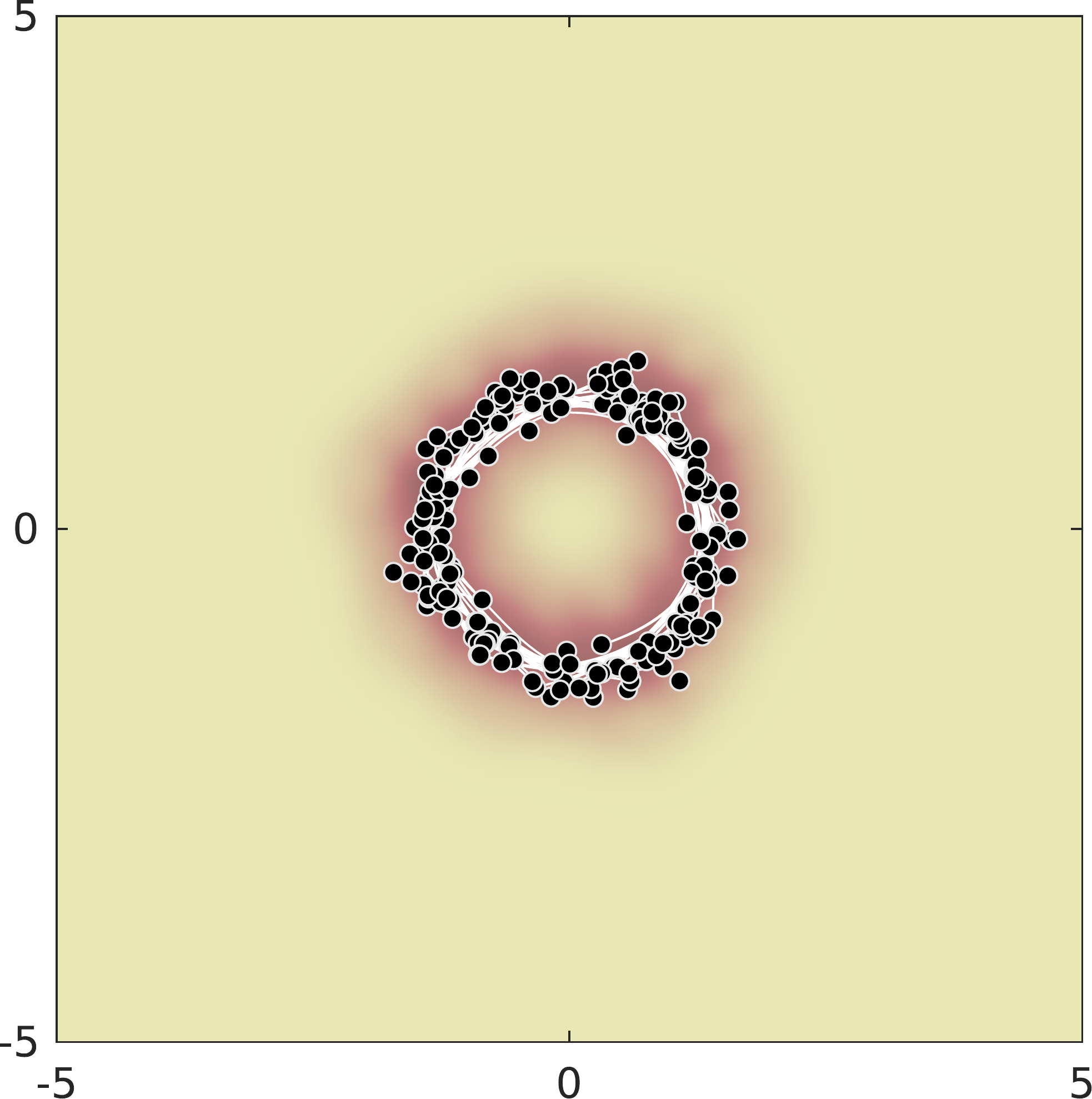}
      \caption{The expected measure of the metric associated with the GP-LVM.
        Notice that the plot is practically identical to Fig.~\ref{fig:big1}d.}
      \label{fig:gplvm}
    \end{wrapfigure}
    The variance of the integral can be expressed as
    \begin{align}
      \var{\bar{h}} &= \E\left[\bar{h}^2\right] - \E\left[\bar{h}\right]^2,
    \end{align}
    where the last term easily is computed from Eq.~\ref{eq:exp_integral}.
    The missing term is
    \begin{align}
      \E\left[\bar{h}^2\right]
        &= \E_{\mat{M}}\left[ \left( \int_{\Omega} h(f(\z)) \sqrt{\det\mat{M}_{\z}} \dif{\z} \right)^2 \right],
    \end{align}
    which generally does not permit a closed-form expression.

    Since geodesics under the expected metric are well-behaved, it is tempting to
    treat the manifold as having this metric.
    From an integration point-of-view this implies working with the measure
    $\sqrt{\det(\E[\mat{M}_{\z}])}$ as suggested by \citet{arvanitidis:iclr:2018}.
    The above analysis, however, indicate that it is perhaps more suitable to use
    the expected measure \eqref{eq:exp_integral}.

    \begin{exbox}
    \begin{example}[a GP prior manifold]\label{ex:gp_prior_manifold_integration}
      There are cases where the choice of measure is of less importance. To see
      this, we return to the Gaussian process prior manifold of Example~\ref{ex:gp_prior_manifold}.
      As before, the metric at $\z$ follows a (scaled) Wishart distribution,
      and
      Theorem~3.2.15 of Muirhead's book \cite{muirhead} tells us that
      \begin{align}
        \E\left[\sqrt{\det(D\mat{M}_{\z})}\right]
          &= \sqrt{2^d \det\bs{\Sigma}_{\z}} \frac{\Gamma\left(\frac{D+1}{2}\right)}{\Gamma\left(\frac{D-d+1}{2}\right)}
        \Leftrightarrow \\
        \E\left[\sqrt{\det(\mat{M}_{\z})}\right]
          &\propto \sqrt{\det\bs{\Sigma}_{\z}}.
      \end{align}
      The measure associated with the expected metric is
        $\sqrt{\det\E\left[\mat{M}_{\z}\right]}
          = \sqrt{\det\bs{\Sigma}_{\z}}$
      and we conclude that for this prior manifold, the two measures are proportional.
    \end{example}
    \end{exbox}

    \begin{exbox}
    \begin{example}[GP-LVM]\label{ex:gplvm_integration}
      Things are not as simple when considering the posterior GP-LVM manifold.
      Here the (scaled) metric at $\z$ follows a non-central Wishart distribution
      \begin{align}
        D\cdot\mat{M} \sim
          \mathcal{W}_d(D, \bs{\Sigma}_{\z}, \bs{\Sigma}_{\z}^{-1}\E[\J]\T \E[\J]).
      \end{align}
      By Theorem~10.3.7 of Muirhead's book \cite{muirhead} we get that
      \begin{align}
      \begin{split}
        \E\Big[&\sqrt{\det(D\mat{M}_{\z})}\Big]
          = \frac{2^{\sfrac{d}{2}}}{\pi D^{\sfrac{d}{2}}}
             \frac{\Gamma\left(\frac{D+2}{4}\right)}{\Gamma\left(\frac{D-d+2}{4}\right)} 
             \sqrt{\det\bs{\Sigma}_{\z}} \\
           & \cdot {_1}F_1\left( -\sfrac{1}{2}, \sfrac{D}{2}, -\sfrac{1}{2} \bs{\Sigma}_{\z}^{-1}\E[\J]\T \E[\J]) \right),
        \label{eq:exp_meas_gplvm}
      \end{split}
      \end{align}
      where ${_1}F_1$ is the confluent hypergeometric function of the first kind.
      On the other hand, the measure associated with the expected metric is
      \begin{align}
        \sqrt{\det\E\left[\mat{M}_{\z}\right]}
          &= \sqrt{\det\left( \sfrac{1}{D}\, \E[\J_{\z}]\T \E[\J_{\z}] +  \bs{\Sigma}_{\z} \right)}
        \label{eq:meas_exp_gplvm}
      \end{align}
      and we see that the two measures appear quite different.
      To understand this difference in practice, we show the volume measure
      of the expected metric \eqref{eq:meas_exp_gplvm} in the background of
      Fig.~\ref{fig:big1}d.
      Similarly, we show the expected volume measure \eqref{eq:exp_meas_gplvm}
      in Fig.~\ref{fig:gplvm}.
      Somewhat surprisingly, there is no visual difference between the two
      different measures. This indicates that for the GP-LVM, the more simple
      volume measure associated with the expected metric may be a good approximation
      to the expected volume measure.
    \end{example}
    \end{exbox}

\section{Deep generative models}\label{sec:deep}
  So far, we have studied kernel based methods due to their ease.
  The key observations, however, generally
  hold true, and we now consider neural networks.

  \subsection{The deterministic setting}
    The natural case is to estimate $f$ with a (potentially deep) feed-forward
    neural network. We call this an \emph{autoencoder} as these classic
    methods are the prime example of such an architecture \cite{rumelhart1985learning},
    though we note that other models such as \emph{generative adversarial networks}
    \cite{goodfellow2014generative} also fall within this category.
    When we consider the associated pull-back metric, then the same considerations
    hold true as in the kernel-based setting (Sec.~\ref{sec:deterministic}).
    That is, if we regularize towards a smooth $f$, then geodesics will naturally
    cross through holes in the data manifold. To validate this, Fig.~\ref{fig:learn_circle_vae}a
    shows that the geodesics of our guiding example are almost straight lines.
    As before, \emph{the lack of uncertainty prevent us from learning the manifold topology and geometry.}

  \subsection{The Bayesian setting}
    In the neural networks literature, (Gaussian) probabilistic mappings $f$
    are commonly represented as \cite{nix:icnn:1994}
    \begin{align}
      f(\z) &= \bs{\mu}(\z) + \text{diag}(\epsilon) \bs{\sigma}(\z),
      \qquad \epsilon \sim \N(\vec{0}, \mat{I}_D),
      \label{eq:deepf}
    \end{align}
    where $\bs{\mu}, \bs{\sigma}: \R^d \rightarrow \R^D $ are neural networks, $\text{diag}(\cdot)$
    constructs a diagonal matrix from a vector, and $\mat{I}_D$ is the $D\!\times\! D$ identity
    matrix.
    This is a key part of variational autoencoders (VAEs) \cite{kingma:iclr:2014, rezende2014stochastic}
    that realize the generative model $\z \sim \N(\vec{0}, \mat{I})$, $\x = f(\z)$.
    %
    From a geometric perspective it is worth noting that the noise $\epsilon$ does not form
    a smooth process. As such, sample paths from
    $f$ are not smooth, and one can question the validity of pull-back metrics
    of this model. If we disregard any such concerns, then it is
    easy to show that \cite{arvanitidis:iclr:2018}
    \begin{align}
      D\cdot\E[\mat{M}_{\z}]
        &= \xtx{ \big( \J_{\z}^{(\bs{\mu})} \big) }
         + \xtx{ \big( \J_{\z}^{(\bs{\sigma})} \big) },
      \label{eq:vae_metric}
    \end{align}
    where $\J_{\z}^{(\bs{\mu})}$ and $\J_{\z}^{(\bs{\sigma})}$ are the Jacobians
    of $\bs{\mu}$ and $\bs{\sigma}$, respectively. As before, the variance of
    the metric also goes to zero when $D \rightarrow \infty$, due to
    the central limit theorem. This can be taken as a hint
    that the expected metric is a reasonable geometric structure for $\Z$.
    
    A perhaps more sensible way to arrive at Eq.~\ref{eq:vae_metric} is to view
    Eq.~\ref{eq:deepf} as a random projection of the deterministic manifold spanned
    by
    \begin{align}
      f: \R^d \rightarrow \R^{2D}, \qquad
      f(\z) &= \left(\begin{array}{c} \bs{\mu}(\z) \\ \bs{\sigma}(\z) \end{array} \right).
      \label{eq:stacked_f}
    \end{align}
    Stacking $\bs{\mu}$ and $\bs{\sigma}$ imply that their Jacobians stack as well,
    such that Eq.~\ref{eq:vae_metric} is the pull-back metric associated with Eq.~\ref{eq:stacked_f}.
    
    \begin{figure*}
    \centering
    \includegraphics[width=0.24\textwidth]{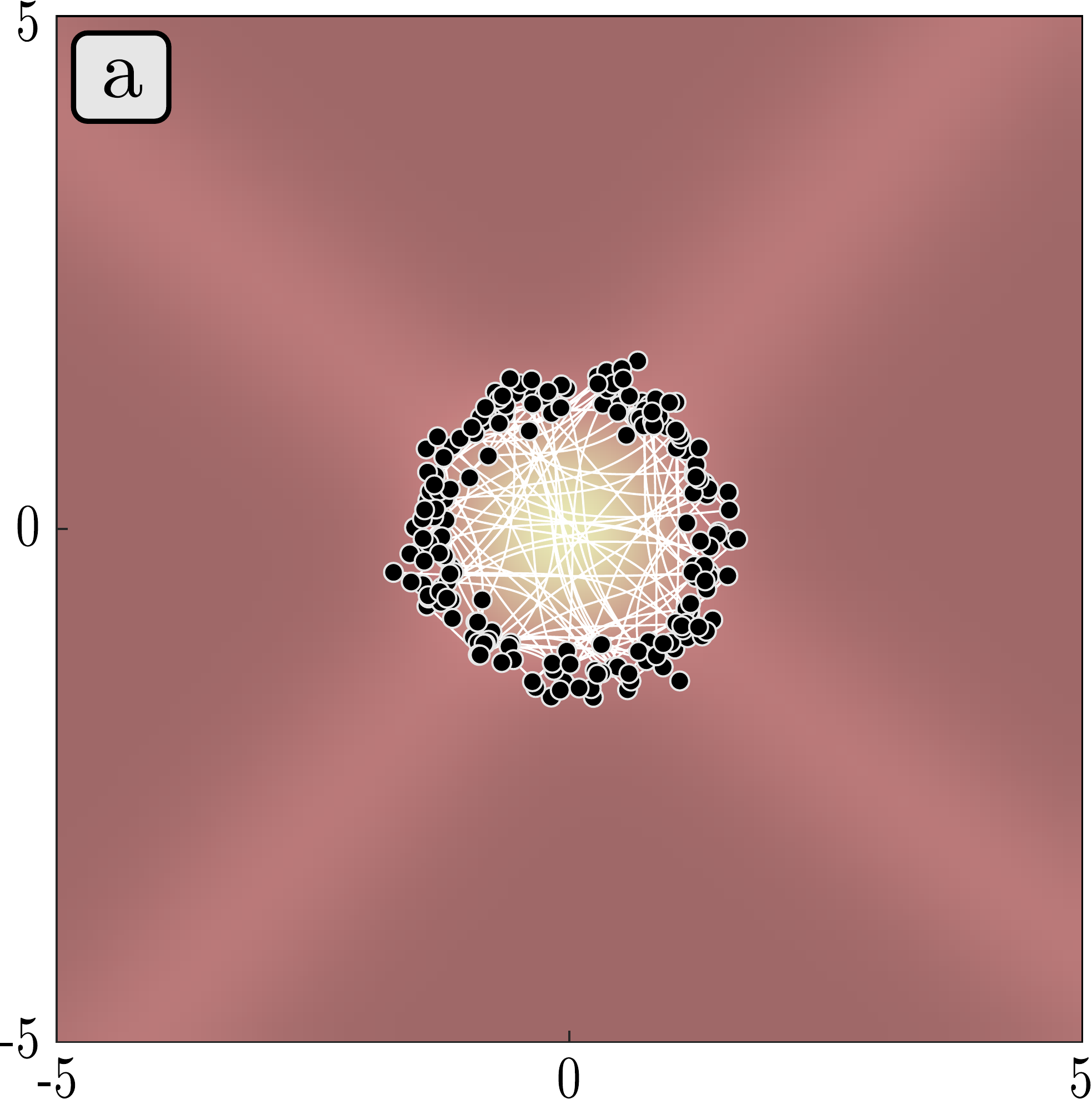}
    \includegraphics[width=0.24\textwidth]{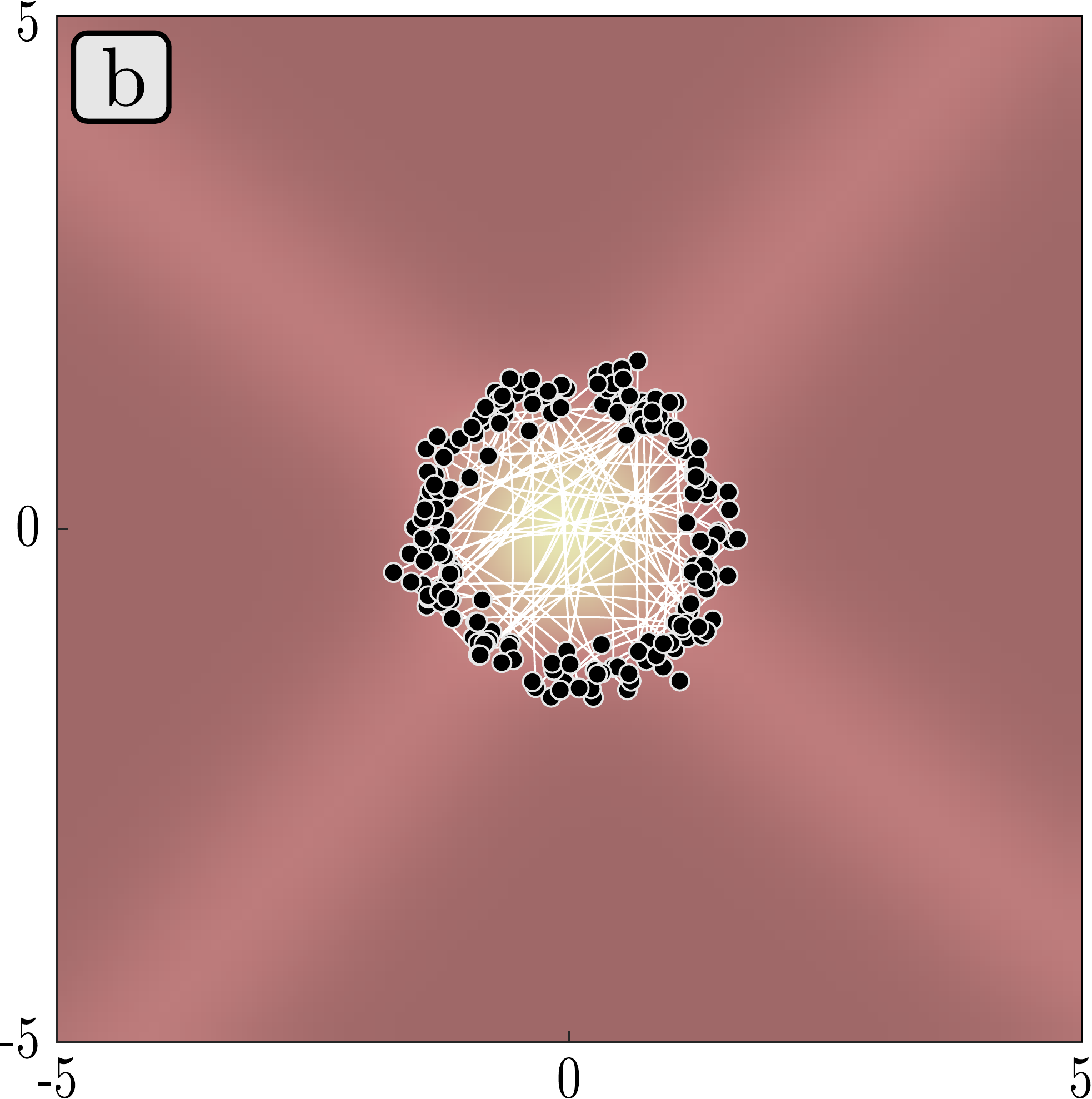}
    \includegraphics[width=0.24\textwidth]{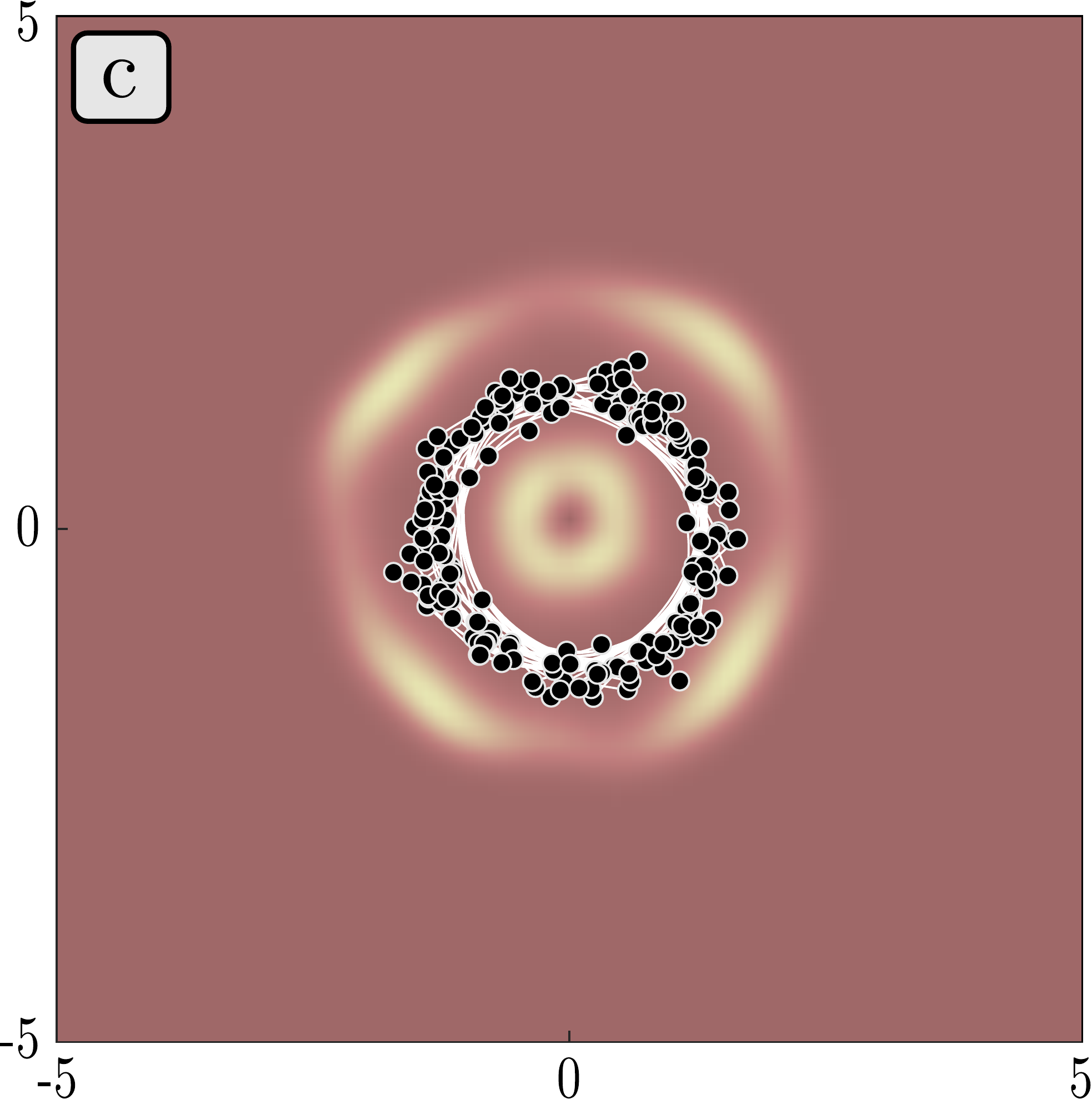}
    \includegraphics[width=0.24\textwidth]{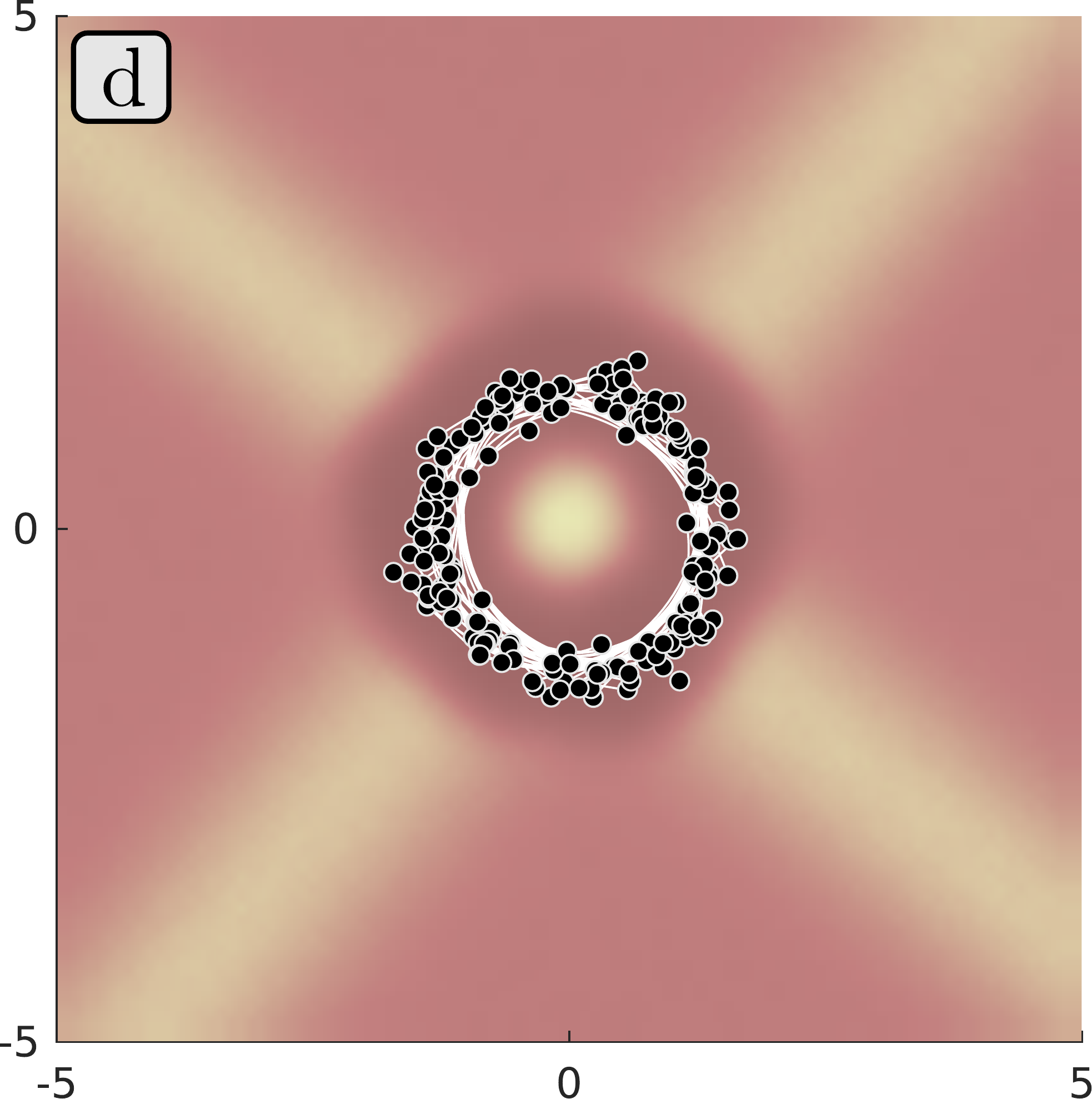}
    \caption{Geodesics for various autoencoders.
      (a) Here $f$ is a smooth feed-forward network.
      (b) A naive VAE where both
          mean and standard deviation of $f$ are smooth feed-forward networks.
      (c) A VAE with decaying precision, \ie the inverse standard deviation of $f$
          is a positive RBF network. 
      (d) Same network as (c) but with background color
      proportional to the expected measure.
      }
    \label{fig:learn_circle_vae}
    \end{figure*}

    Returning to our guiding example, we let
    $\bs{\mu}$ and $\bs{\sigma}$ be smooth feed-forward neural networks.
    Figure~\ref{fig:learn_circle_vae}b shows that recovered geodesics
    are almost straight lines, \ie the model failed to capture the
    data geometry. This is because $\sigma$ is a poor
    proxy for uncertainty \cite{arvanitidis:iclr:2018}. When $\sigma$
    is a feed-forward neural network, we assume that we can smoothly
    interpolate the uncertainty estimates recovered at $\z_{1:N}$.
    But smooth interpolation of uncertainty is nonsensical.\footnote{Consider
      two low-variance temperature readings at the poles of our planet. If we,
      from this data, interpolate the temperature at the equator, then a smooth
      interpolation of the uncertainty would imply a very certain prediction.}
    To counter this, \citet{arvanitidis:iclr:2018} model
    $\bs{\sigma}^{-1}$ with a positive RBF network \cite{que:aistats:2016}, which
    ensure that uncertainty grows away from the data, and
    Fig.~\ref{fig:learn_circle_vae}c
    shows that it allows us to recover the geometry of the data manifold.
    
    In Example~\ref{ex:gplvm_integration}, we saw that for the GP-LVM the measure associated with
    the expected metric was practically identical to the expected measure, even
    if their mathematical expressions are quite different. This does not appear
    to be the case for the variational autoencoder. Figure~\ref{fig:learn_circle_vae}c
    shows the measure associated with the expected metric, Fig.~\ref{fig:learn_circle_vae}d
    shows the expected measure (computed using sampling).
    Unlike for the GP-LVM, we now see significant differences between the two measures.
    At this stage, it is unclear which measure is to be preferred from a practical
    perspective. We see that both measures exhibit a somewhat arbitrary
    behavior, which we take as a hint that the RBF network for inverse variance
    does not provide an excellent fit to the data.

\section{Quantitative summary}
  \begin{wraptable}[11]{r}{0.35\textwidth}
    \vspace{-4mm}
    \resizebox{0.35\textwidth}{!}{
    \begin{tabular}{l|cc}
      \hline
      \textbf{Method} & \textbf{corr} & \textbf{Hausdorff} \\ \hline
      GP-LVM          & 0.996819      & 0.858773 \\
      KRM             & 0.843259      & 5.443746 \\
      KRM+ridge       & 0.892995      & 3.467452 \\
      AE              & 0.974833      & 2.825305 \\
      VAE             & 0.975200      & 2.826954 \\
      RBF VAE         & 0.982504      & 0.788519 \\
      \hline
    \end{tabular}}
    \vspace{-2mm}
    \caption{Correlation between length of true and estimated geodesics, and
             Hausdorff distances between these curves.}
    \label{tab:numbers}
  \end{wraptable}
  Throughout the previous section we have used a simple example to illustrate
  the fundamental bias in deterministic geometry estimation. For completeness,
  we here quantify these results, but emphasize that our main objective is to
  understand this bias, rather than to perform empirical studies.
  Since we know that ground truth geodesics are circular arcs, we can compare
  these to estimated geodesics. As a first measure
  of quality, we compute the correlation between the estimated geodesic lengths
  and the length of the ground truth geodesics. Table~\ref{tab:numbers}
  show this correlation for each considered model, and individual correlation
  plots are found in Appendix~\ref{app:corr}. We observe that the GP-LVM
  achieve an almost perfect correlation, while deterministic kernel methods fare significantly
  worse. Individual correlation plots show that for deterministic models
  short curves provide a better correlation than long curves, which is in line
  with the Riemannian assumption of a locally Euclidean model. All autoencoder-based
  models achieve a strong correlation, which is somewhat surprising given the
  almost straight geodesics found by deterministic methods.
  
  As a second quality measure, we report the average Hausdorff distance \cite{rockafellar2009variational}
  between estimated and ground truth geodesics (Table~\ref{tab:numbers}).
  We see that the GP-LVM and the VAE with RBF variance
  estimation significantly outperform the other methods. This match
  the visual observations made throughout the paper.
  
\section{Previous work}
  Pull-back metrics have been studied in mathematics at least since the seminal
  work of Gauss \cite{gauss:surfaces:1827}, and formed the initial foundation of
  Riemannian geometry. In machine learning, these metrics have
  only been studied in few instances. 
  \citet{Tosi:UAI:2014} was the first to give the latent space of the GP-LVM \cite{gplvm}
  a geometric foundation. \citet{bishop1997magnification} used the deterministic
  volume measure of a \emph{Generative Topographic Map (GTM)} \cite{bishop1998gtm} as a
  visualization tool; our analysis imply that incorporating uncertainty
  should improve such a tool.
  Recently, several authors have studied the geometry of deep generative models
  \cite{shao2017riemannian, chen2018metrics, laine2018feature-based, arvanitidis:iclr:2018}.
  \citet{shao2017riemannian} and \citet{chen2018metrics} consider pull-back metrics
  of VAEs, but only
  consider the mean of $f$; in our terminology they therefore consider
  autoencoders rather than variational autoencoders.
  \citeauthor{shao2017riemannian} note that most geodesics in their model are straight lines and speculate
  that this is because most data manifold are actually flat. Our analysis shows
  that this conclusion is most likely incorrect, and that flatness is an
  artifact of disregarding uncertainty. 
  \citet{arvanitidis:iclr:2018} also considered pull-back metrics of variational
  autoencoders by taking the expected metric. Here significant curvature is reported,
  which coincides with intuition. 
  
  As an alternative to estimating a geometry that also reflect topology, one can
  bias the estimated geodesics such that they are attracted to data. \citet{chen2018metrics}
  initialize geodesics by a density maximizing curve before optimizing curve energy,
  and in later work \citet{chen2018fast} force geodesics to be polygonal curves
  that interpolate the data. From our perspective, these approaches work
  around a deeper more fundamental problem, as it is not clear if these biased
  curves correspond to geodesics under any metric (they most surely do not minimize
  the energy associated with the deterministic pull-back metric).

\section{Concluding remarks}
  The driving motivation for introducing pull-back metrics in the latent space
  of a generative model is to avoid arbitrariness in parametrizing
  the latent space. This is an important issue if we are to interpret the latent
  variables of a fitted model. We have argued that geometry provides \emph{a} solution
  to the issue, but emphasize that this need not be the only solution.
  We have demonstrated that methods that do not quantify their uncertainty
  cannot, in a meaningful way, capture the geometry of a data manifold. The key issue
  is that the usual smoothness assumptions imply that holes in the data manifold
  are interpolated so smoothly that geodesics are encouraged to pass through the
  holes rather than stay on the manifold. Methods that provide reasonable estimates
  of the uncertainty of the estimated manifold naturally avoid this issue as
  the uncertainty directly alters the estimated geometry. We find that uncertainty quantification
  in manifold learning ends up playing the role of topology in classic geometry:
  uncertainty informs us about holes and boundaries in the manifold and provides
  us with a global notion of connectivity. Disregarding uncertainty, thus, imply
  disregarding the most fundamental aspects manifold learning.
  
  We have provided an extensive analysis of the geometry of Gaussian
  latent variable models, and have developed an elementary theory for stochastic
  Riemannian manifolds. This appear to be the first of its kind.
  We have seen that minimizing expected
  curve energy on a stochastic manifold does not imply minimization of expected
  curve length, which forces us to reconsider which measure define the most
  natural interpolants. A more formal treatment of this material alongside
  approximation bounds when using expected metrics have recently appeared \cite{eklund2019expected}.
  
  Parts of our analysis can be extended to models based on neural networks.
  This raises two key issues for future research: 1) since sample paths from
  deep generative models are not continuous it is perhaps not a good idea
  to enforce a geometric analysis. It then becomes interesting to determine if
  such smoothness can be introduced in deep generative models without
  sacrificing the computational efficiency of the models. 2) Uncertainty
  is essential for estimating the geometric structure of a data manifold,
  but current deep generative models provide rather poor
  estimators of uncertainty. A heuristic from \citet{arvanitidis:iclr:2018}
  seems to work, but more principled methods would
  be valuable; \citet{detlefsen2019reliable} provide some early work, but the
  question largely remain open.
  
  Finally, we repeat the key point of the paper:
  \emph{without uncertainty quantification, we cannot learn the geometric
  structure of a data manifold, and any attempt to do so is bound to fail beyond
  the most simple examples.}
  
\subsubsection*{Acknowledgments}
  The author is grateful to Vagn Lundsgaard Hansen, Martin J{\o}rgensen,
  Georgios Arvanitidis, Lars Kai Hansen, David Eklund and Aasa Feragen for enlightening discussions.
  SH was supported by a research grant (15334) from VILLUM FONDEN. This project has
  received funding from the European Research Council (ERC) under the European
  Union's Horizon 2020 research and innovation programme (grant agreement n\textsuperscript{o} 757360). 

{\small
\bibliographystyle{abbrvnat}
\bibliography{paper}
}

\newpage
\appendix
\section{Experimental details}\label{app:data}
  \paragraph{Data creation:}
    We sample $N = 200$ numbers $\{t_n \}_{n=1}^N$ uniformly over the interval
    $[0, 2\pi]$, and embed these first in $\R^3$ as
    \begin{align}
      \vec{v}_n &= 
      \begin{bmatrix}
        \cos(t_n) \\ \sin(t_n) \\ \cos(t_n) \sin(t_n)
      \end{bmatrix}
    \end{align}
    The data is then generated as
    \begin{align}
      \x_n &=
      \begin{bmatrix}
        \vec{v}_n \\
        \vec{0}
      \end{bmatrix}
      +
      \sigma\bs{\epsilon},
    \end{align}
    where $\vec{0}$ is a $D\!-\!3$ dimensional vector of zeros, $\bs{\epsilon}$
    is a $D$ dimensional vector drawn from a unit Gaussian, and $\sigma = 0.1$
    captures ``off manifold''-noise. Here we embed into a $D=1000$ dimensional
    vector space.
    
    Since the data fundamentally live on a unit circle (which we then nonlinearly
    embed in $\R^D$), we fix the latent variables $\z_n$ to be the first two
    dimensions of $\x_n$; this correspond to points on the unit circle with
    added Gaussian noise. In order to compare different models, we fix the latent
    variables to have the same values in all models.
  
  \paragraph{Kernel methods:}
    We first fit a model using Gaussian process (GP) regression from $\z_{1:N}$ to $\x_{1:N}$
    with hyperparameters estimated using maximum likelihood. We do not update
    the latent variables $\z_{1:N}$ as is commonly done for the GP-LVM as this
    would complicate a direct comparison between different models. The methods
    based on kernel ridge regression are simple taken as the GP mean function.
    This ensure that the exact same hyperparameters are used in the GP and
    kernel ridge regression experiments. Again, this choice was made to simplify
    the comparison of different methods.
  
  \paragraph{Neural network methods:}
    We first train a multilayer perceptron from $\z_{1:N}$ to $\x_{1:N}$ according
    to the usual autoencoding criterion. We use this to form our autoencoder models.
    We keep this mapping as the mean function $\mu$ of the variational autoencoders (VAE),
    and fit the uncertainty $\sigma$ according to the usual VAE criterion.
    As before, we take this restricted approach to ensure that models are as
    comparable as possible.

\section{Correlation plots}\label{app:corr}
\begin{figure*}
  \begin{tabular}{ccc}
  \tiny{Kernel regression with Gaussian kernel} &
  \tiny{Kernel regression with Gaussian+linear kernel} &
  \tiny{GP-LVM} \\
  \includegraphics[width=0.3\textwidth]{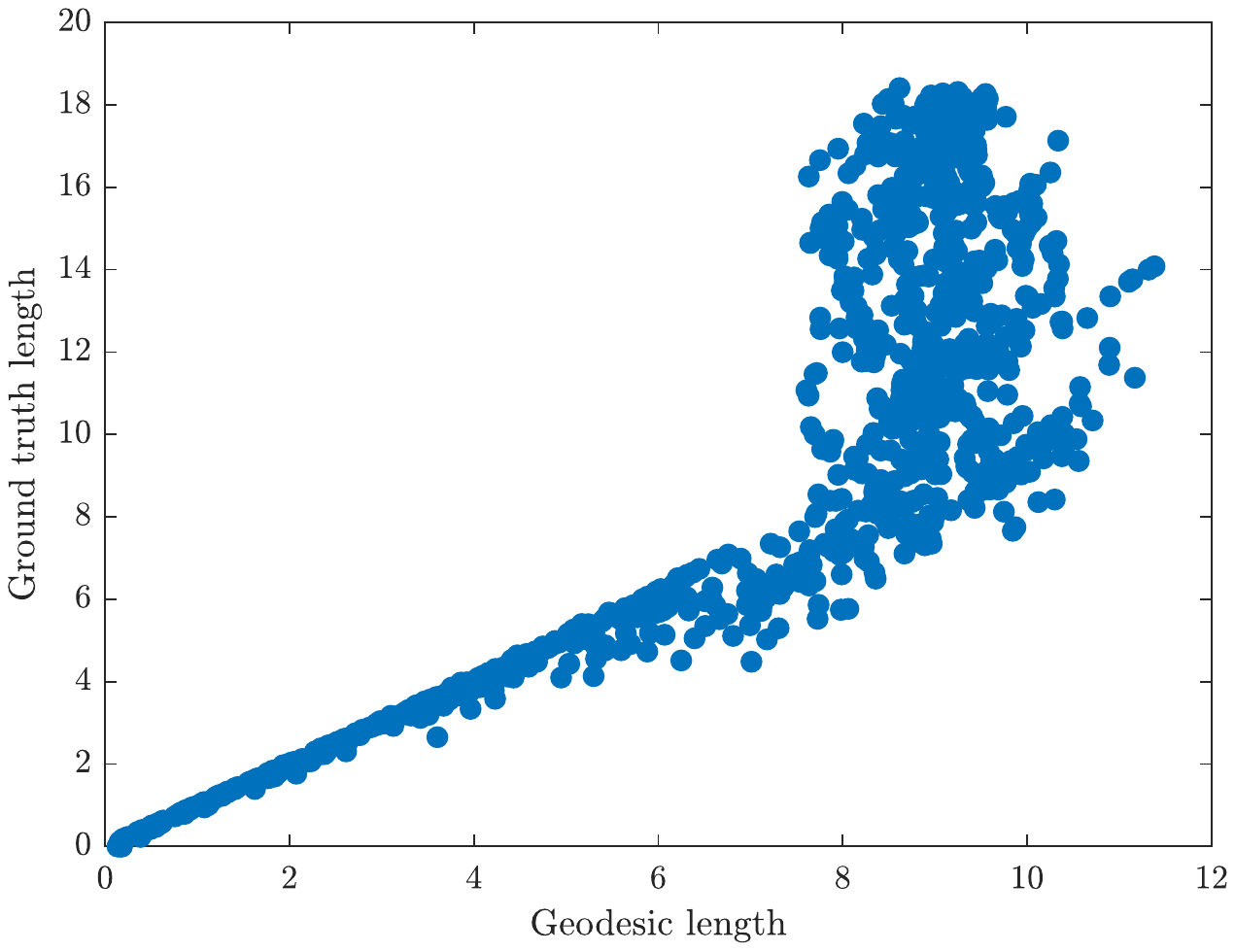}          &
  \includegraphics[width=0.3\textwidth]{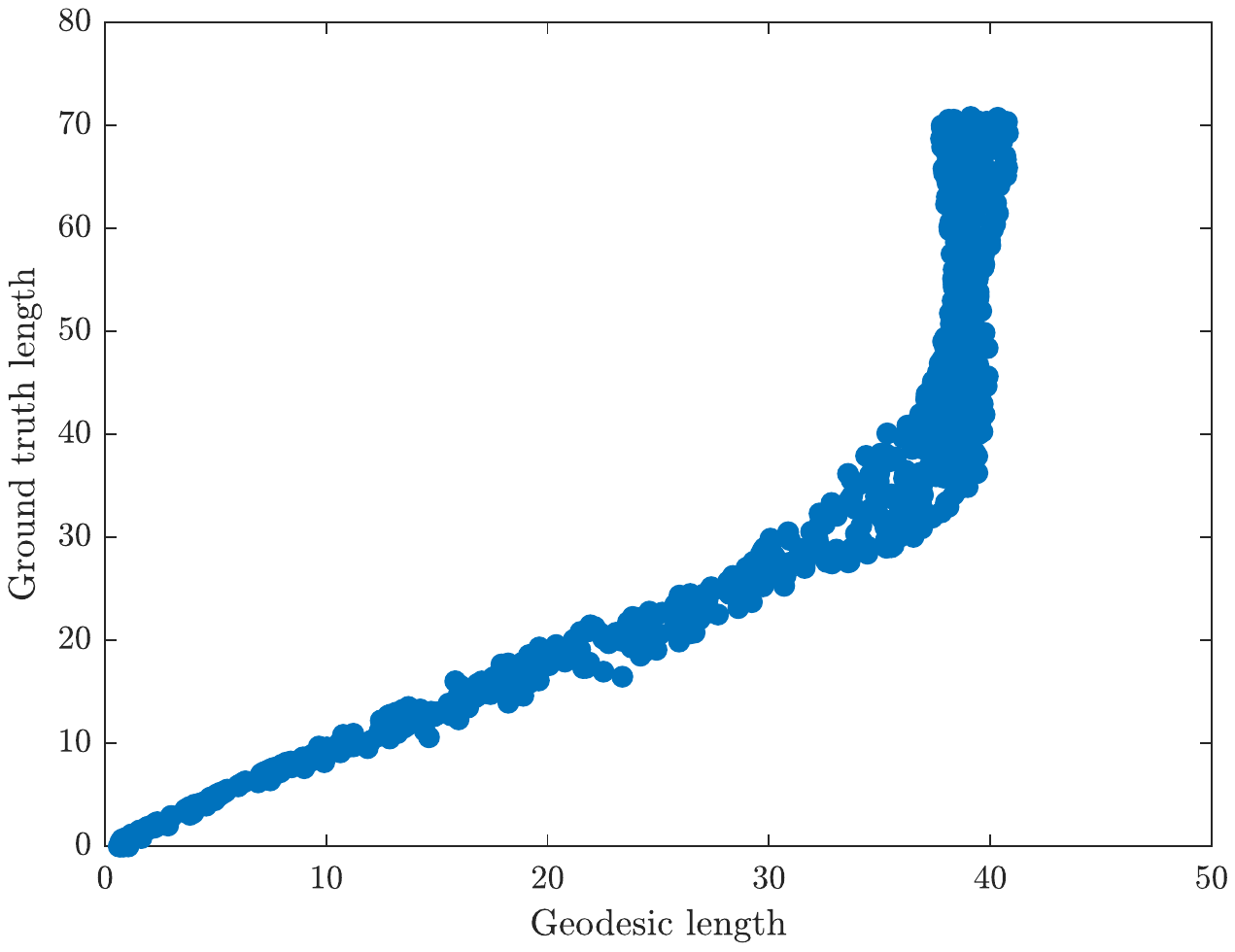}    &
  \includegraphics[width=0.3\textwidth]{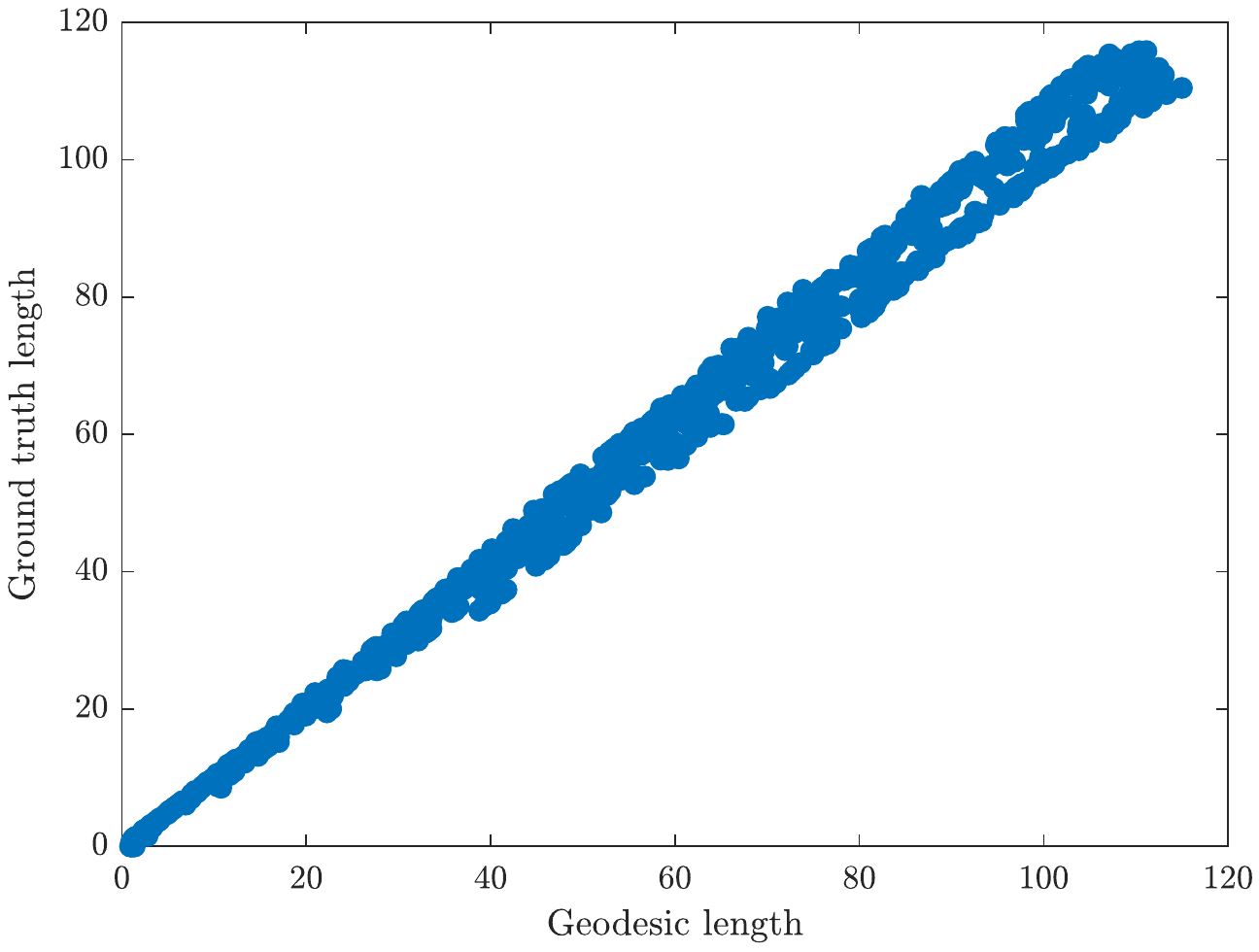}        \\
  & & \\
  \tiny{Autoencoder} &
  \tiny{Variational autoencoder} &
  \tiny{Variational autoencoder with RBF precision} \\
  \includegraphics[width=0.3\textwidth]{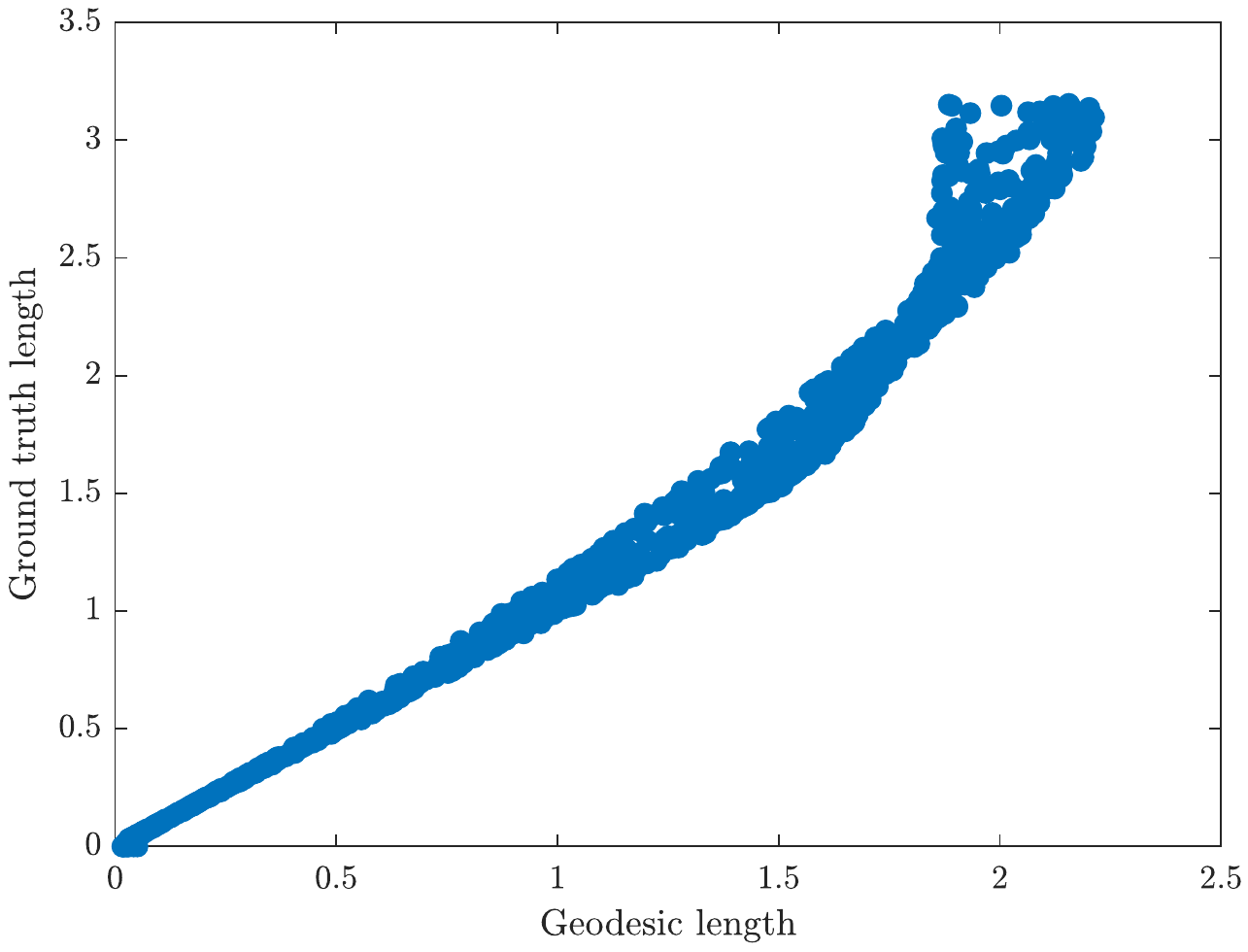}           &
  \includegraphics[width=0.3\textwidth]{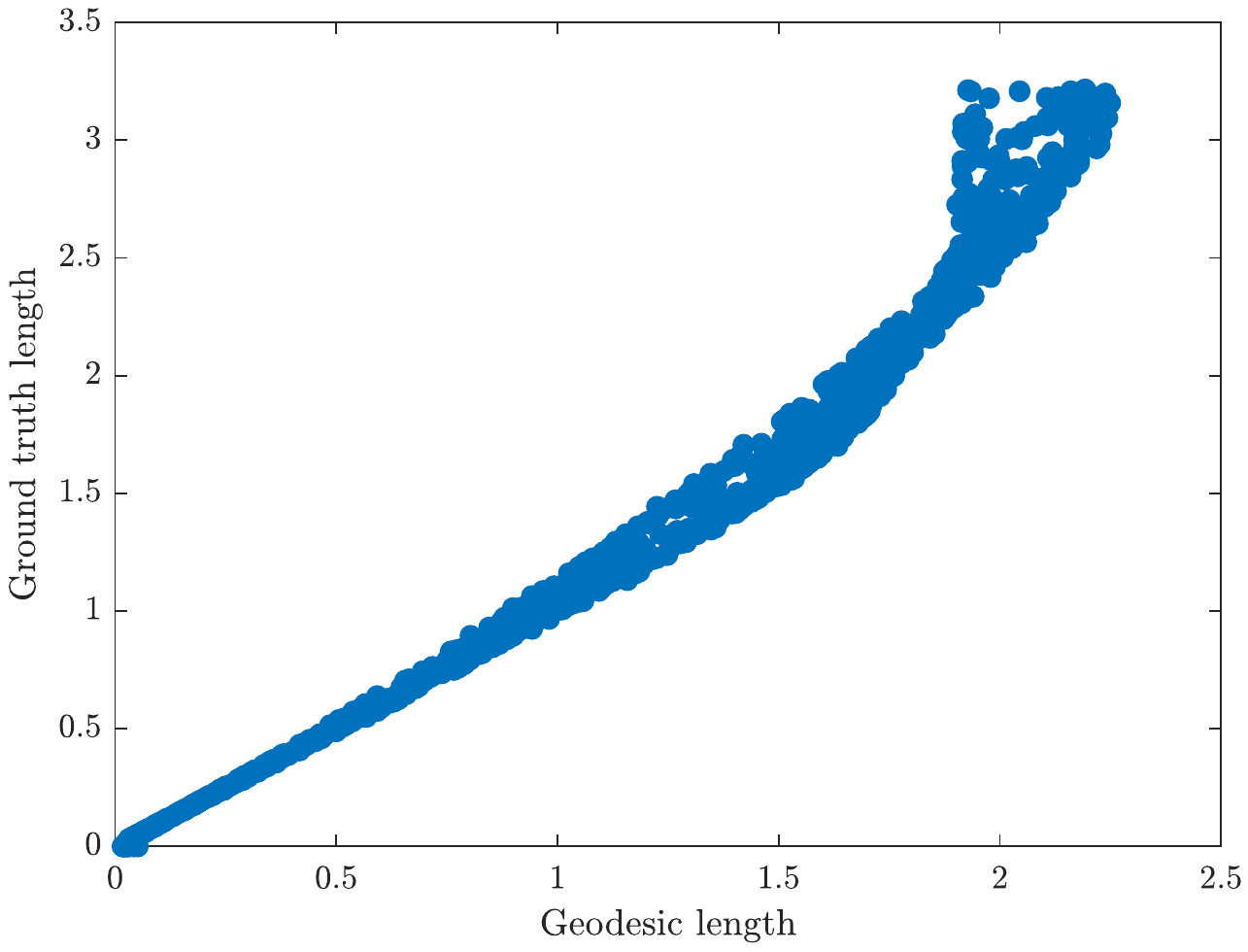}          &
  \includegraphics[width=0.3\textwidth]{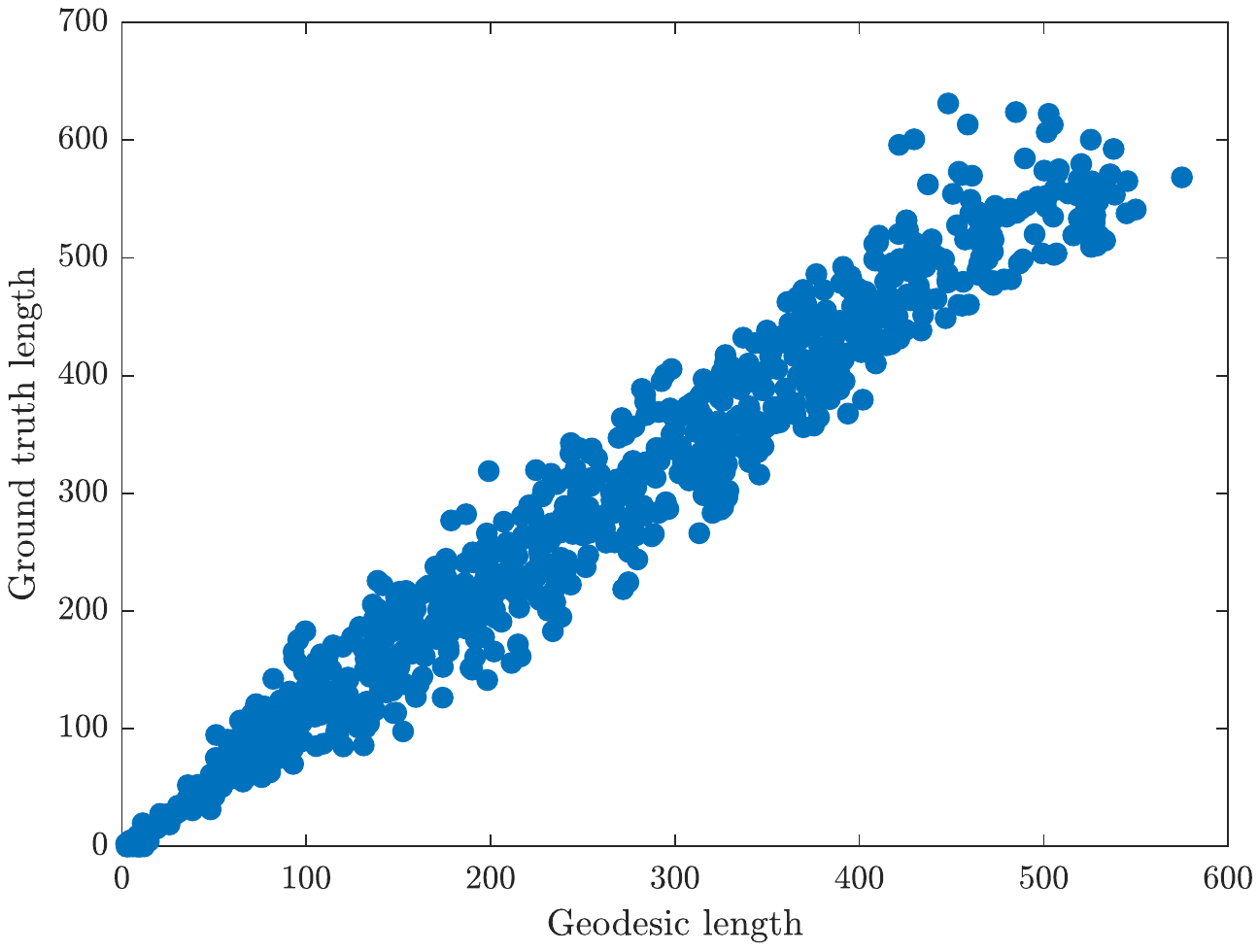}      
  \end{tabular}
  \caption{Correlation between length of ground truth geodesics and estimated
    geodesics for different models.}
  \label{fig:corr}
  \end{figure*}

  In the main text we quantitatively compare models in two ways. The key idea is to
  take advantage of the fact that we know that ground truth geodesics should be
  circular arcs. Let $\vec{c}_{\text{gt}}$ denote a ground truth geodesic, and
  let $\vec{c}_{\text{est}}$ denote an estimated geodesic. As a first measure
  of quality, we compute the length of each curve under the model-specific metric
  \begin{align}
    \mathcal{L}_{\text{gt}} = \mathcal{L}(\vec{c}_{\text{gt}})
      &= \int_a^b \| \partial_t \vec{c}_{\text{gt}} \|_{\mat{M}} \dif{t} \\  
    \mathcal{L}_{\text{est}} = \mathcal{L}(\vec{c}_{\text{est}})
      &= \int_a^b \| \partial_t \vec{c}_{\text{est}} \|_{\mat{M}} \dif{t}.
  \end{align}
  Both integrals are evaluated using standard quadrature. In the main text, we
  report the correlation between $\mathcal{L}_{\text{gt}}$ and $\mathcal{L}_{\text{est}}$
  for randomly sampled points. More insight into the empirical behavior of the
  different models can be found be directly plotting $\mathcal{L}_{\text{gt}}$
  and $\mathcal{L}_{\text{est}}$ against each other, which we do in Fig.~\ref{fig:corr}.
  Here we see a clear trend that methods with no or meaningless uncertainty
  quantification all exhibit a trend that short curves correlate well with
  the ground truth, but long curves do not. As the Riemannian model is that
  we work with locally Euclidean models, this behavior is not too surprising.
  We see that with methods with reasonably sensible uncertainty quantification
  the two lengths have a more well-behaved correlation. It is worth pointing
  out that that VAE with RBF precision is not as well behaved as the GP-LVM, 
  which indicate that the variance estimation leaves something to be desired.
  
  As a second measure of quality we consider the Haussdorf distance between
  $\vec{c}_{\text{gt}}$ and $\vec{c}_{\text{est}}$. This is defined as
  \begin{align}
  \begin{split}
    \text{dist}_H(\vec{c}_{\text{gt}}, \vec{c}_{\text{est}})
      = \max\Huge\{
           \sup_{x \in \vec{c}_{\text{gt}}} \inf_{y \in \vec{c}_{\text{est}}} \| x - y \|, \,
           \sup_{y \in \vec{c}_{\text{est}}} \inf_{x \in \vec{c}_{\text{gt}}} \| x - y \|
         \Huge\}.
  \end{split}
  \end{align}
  We refer to the main text for results and discussion.
  
\end{document}